%% file: iclr2027_conference.tex
\documentclass{article} 
\usepackage{iclr2027_conference,times}

\input{math_commands.tex}

\usepackage{hyperref}
\usepackage{url}

\usepackage{microtype}
\usepackage{graphicx}
\usepackage{capt-of}
\usepackage{subfigure}
\usepackage{booktabs} 

\usepackage{algorithm}
\usepackage{algorithmic}

\usepackage{amssymb}
\usepackage{mathtools}
\usepackage{amsthm}

\usepackage[capitalize,noabbrev]{cleveref}

\usepackage{setspace}
\usepackage{cancel}
\theoremstyle{plain}

\theoremstyle{definition}

\theoremstyle{remark}

\usepackage[textsize=tiny]{todonotes}

\usepackage{multirow}
\usepackage{makecell}
\usepackage{enumitem}
\usepackage{makecell}
\usepackage{colortbl}
\usepackage{wrapfig}    
\usepackage[table]{xcolor} 

\definecolor{yellowavg}{HTML}{FFF9E6}
\definecolor{blueavg}{HTML}{E6F2F7}

\newcommand{\cmark}{\textcolor{green!60!black}{\checkmark}}
\newcommand{\xmark}{\textcolor{red!70!black}{$\bm{\times}$}}

\newcommand{\ul}{\underline}

\title{MOER: Mix-Policy Optimization under Expert Prefix Guidance with Selective Rollback}

\author{Xuyang Zhong$^{1}$ \quad Jiuding Yang$^{2}$ \quad Yiwen Guo$^{3,*}$  \quad Chen Liu$^{1,*}$\\
$^{1}$City University of Hong Kong  \quad $^{2}$LIGHTSPEED  \quad $^{3}$Independent Researcher\\
$^{*}$Corresponding Author
}

\iclrfinalcopy 
\begin{document}

\maketitle

\begin{abstract}
Mix-policy optimization incorporates off-policy expert data into the reinforcement learning (RL) loop to transcend the capability boundary of on-policy RL. However, existing methods either provide static full-trace guidance that suppresses self-exploration, or adopt expert prefix guidance without correctness guarantees, leading to suboptimal performance or even training collapse. In this paper, we propose \textbf{MOER}, a mix-policy optimization framework that harmonizes expert guidance with self-exploration. MOER injects a decaying expert chain-of-thought prefix into the prompt and lets the model complete the remaining reasoning, with the prefix length progressively annealed to encourage self-exploration. A selective rollback mechanism replaces failed completions with the original full expert trace, guaranteeing the correctness of every off-policy trajectory. Furthermore, a harmonized mix-policy objective optimizes on-policy tokens with a GRPO loss while supervising both expert prefix and completion tokens with a trust-region SFT loss, substantially stabilizing training. Extensive experiments show that MOER achieves the highest average scores in mathematical and general domain reasoning tasks, outperforming all basic and mix-policy baselines, and generalizes consistently across diverse tasks, model scales, and base model categories. Codes are available at \href{https://github.com/CityU-MLO/MOER}{https://github.com/CityU-MLO/MOER}.
\end{abstract}

\input{chapter/intro}
\input{chapter/related}

\input{chapter/method}
\input{chapter/exp}
\input{chapter/conclusion}

\bibliography{iclr2027_conference}
\bibliographystyle{iclr2027_conference}

\newpage
\appendix
\input{appendix/analysis}

\input{appendix/exp}

\end{document}

%% file: math_commands.tex
\usepackage{amsmath,amsfonts,bm}

\def\eqref#1{equation~\ref{#1}}

\def\1{\bm{1}}

\def\vx{{\bm{x}}}
\def\vy{{\bm{y}}}

\DeclareMathAlphabet{\mathsfit}{\encodingdefault}{\sfdefault}{m}{sl}
\SetMathAlphabet{\mathsfit}{bold}{\encodingdefault}{\sfdefault}{bx}{n}

\def\gB{{\mathcal{B}}}

\def\gD{{\mathcal{D}}}

\def\gG{{\mathcal{G}}}

\def\gL{{\mathcal{L}}}

\def\gY{{\mathcal{Y}}}



%% file: chapter/intro.tex
\section{Introduction}

Reinforcement learning with verifiable rewards (RLVR) has recently emerged as a powerful paradigm for eliciting complex reasoning capabilities in large language models (LLMs), as exemplified by breakthroughs such as DeepSeek-R1 \citep{guo2025deepseek}. Built upon group-based policy optimization algorithms such as GRPO \citep{shao2024deepseekmath}, RLVR enables models to develop long chain-of-thought (CoT) reasoning and sophisticated behaviors including self-reflection and self-correction, driven purely by outcome-based reward signals.

Nevertheless, purely on-policy reinforcement learning (RL) is inherently bounded by the knowledge already present in the base model: unlike supervised fine-tuning (SFT), which acquires genuinely new reasoning patterns, this approach amplifies existing ones. Consequently, models without robust foundational capabilities rapidly reach a performance plateau \citep{yue2026does}. To overcome this limitation, a growing line of work incorporates off-policy expert data into the RL loop, giving rise to \textit{the mix-policy paradigm}. Pioneering methods such as LUFFY \citep{yan2026learning} inject complete expert traces into on-policy rollout groups, while ReLIFT \citep{ma2026learning} and CHORD \citep{zhang2026onpolicy} interleave or reweight SFT-style objectives alongside RL updates. Despite their promise, these approaches expose the model to expert supervision in a static and indiscriminate manner throughout training, which gradually suppresses self-exploration, prevents the model from internalizing reasoning behaviors beyond imitation.

An alternative and seemingly appealing direction is expert-guided generation, where a prefix of the expert CoT is provided as a hint and the model completes the remaining reasoning steps \citep{huang2026blending,su2026trustregion}. However, two critical deficiencies remain unresolved. First, the correctness of model completions conditioned on expert prefixes is not guaranteed: failed completions contaminate the off-policy supervision signal, which either degenerates the method into plain on-policy RL \citep{huang2026blending} or triggers training collapse with exploding response lengths when the off-policy guidance is overly aggressive \citep{su2026trustregion}. Second, existing formulations treat the completion tokens as on-policy tokens and optimize them with the standard clipped surrogate objective. 
However, prefix conditioning drives the model into context states rarely reached by the current policy when conditioned solely on the original prompts, effectively rendering it an off-policy rollout.
Overlooking this mismatch leads to training instability.

\begin{figure}[t]
    \centering
    \includegraphics[width=\linewidth]{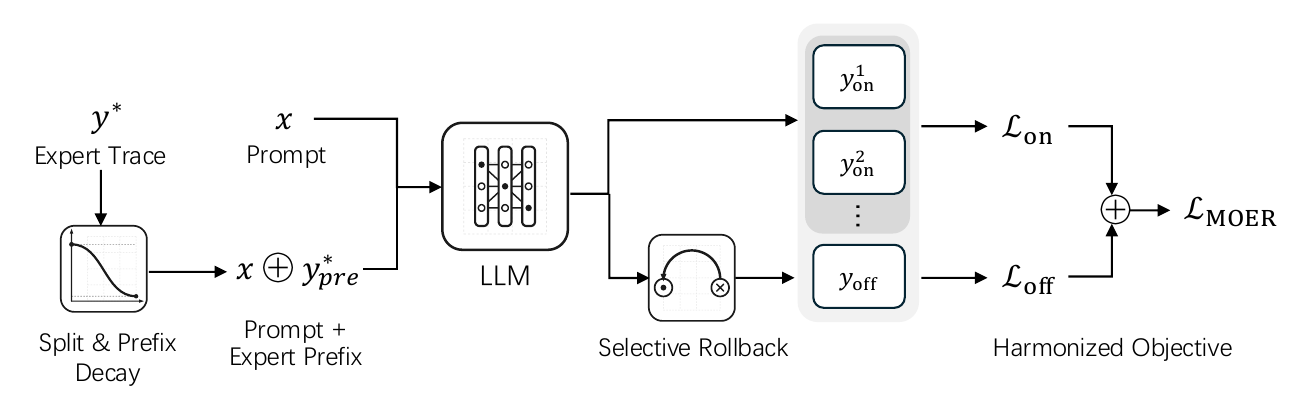}
    \vspace{-1em}
    \caption{Overview of\textbf{ MOER}, which incorporates decaying expert prefix guidance with a selective rollback mechanism and harmonized mix-policy objective. Decaying expert prefix guidance encourages the model's self-exploration capability as the training progresses. Selective rollback mechanism guarantees the correctness of off-policy traces. Harmonized mix-policy objective makes the training stable and effective.}
    \label{fig:overview}
    \vspace{-0.5em}
\end{figure}

To address these challenges, we propose \textbf{M}ix-policy \textbf{O}ptimization under \textbf{E}xpert prefix guidance with selective \textbf{R}ollback (\textbf{MOER}),
as illustrated in Figure \ref{fig:overview}. MOER is built upon three key designs. \textbf{(1) Decaying expert prefix guidance:} instead of injecting the full expert trace or its prefix of a fixed length, we add a truncated expert CoT whose length is progressively annealed following a cosine schedule and let the model complete the remaining reasoning.
Inspired by curriculum learning \citep{bengio2009curriculum}, this design shifts the learning focus from imitation towards self-exploration as training progresses. \textbf{(2) Selective rollback:} whenever a completion conditioned on the expert prefix fails verification, we seamlessly roll back to the concatenation of the original prompt and the full expert trace at no additional computational cost, thereby guaranteeing the correctness of every off-policy trajectory. \textbf{(3) Harmonized mix-policy objective:} we optimize on-policy tokens with a clipped GRPO objective while treating both expert prefix tokens and completion tokens as off-policy tokens supervised by a trust-region objective, which yields smoother gradient weights and effectively stabilizes training.

Extensive experiments demonstrate the effectiveness of MOER. On Qwen2.5-Math-7B, MOER achieves an average score of \textbf{49.1} across six mathematical reasoning benchmarks and \textbf{59.0} on three out-of-distribution general reasoning benchmarks, outperforming all baselines, including LUFFY and ReLIFT. The improvements generalize across different tasks, model scales, and base model categories: MOER significantly outperforms LUFFY on coding task and attains the best overall results on Qwen2.5-Math-1.5B, Qwen2.5-Instruct-7B, and Qwen2.5-Instruct-14B. In-depth analyses further verify that each component of MOER contributes indispensably to training stability and final performance. By contrast, existing prefix-guided baselines either collapse early or degenerate into plain GRPO.

Our contributions can be summarized as:
\begin{itemize}[leftmargin=1.5em,itemsep=0.1em,topsep=0.2em]
    \item We propose \textbf{MOER}, a mix-policy optimization framework that incorporates decaying expert prefix guidance with a selective rollback mechanism, achieving a dynamic equilibrium between expert imitation and self-exploration while guaranteeing the correctness of off-policy trajectories.
    \item We identify that expert prefix guidance induces an off-policy shift of the output distribution and, together with correctness filtering, a success-conditioned selection bias on completion tokens. Accordingly, we design a harmonized mix-policy objective, which optimizes on- and off-policy tokens with distinct loss formulations to improve training stability.
    \item We empirically demonstrate that MOER establishes new state-of-the-art results among mix-policy methods on Qwen2.5-Math-7B, with an average gain of +0.5 points on six math benchmarks and +1.6 points on out-of-distribution general reasoning tasks over the strongest baselines. The performance improvement generalizes consistently across diverse tasks, model scales, and base model categories.
\end{itemize}

%% file: chapter/related.tex
\section{Related Work}
Reinforcement learning (RL) has recently driven major advances in LLM reasoning \citep{liu2025understanding,hu2026open}. A rich line of work has refined RL objectives, from PPO's clipped trust region \citep{schulman2017proximal} and GRPO's group-based advantage estimation~\citep{shao2024deepseekmath} to variants like DAPO~\citep{yu2026dapo} and Dr.~GRPO~\citep{liu2025understanding}. Despite these advances, a fundamental challenge persists: RL-based exploration is often constrained by the model's initial knowledge, making it difficult to discover superior reasoning pathways~\citep{yue2026does}. To address this, recent works shift focus towards incorporating off-policy expert data into the on-policy RL loop, moving beyond the serial SFT-then-RL paradigm. We introduce these works in various categories as follows.

In \textbf{direct data mixing} methods, expert trajectories are combined with self-rollout generations at the data level. SimpleMix~\citep{li2025simplemix} combines off-policy and on-policy data via simple dataset-level sampling within a DPO framework. LUFFY~\citep{yan2026learning} incorporates off-policy expert trajectories directly into the on-policy rollout groups within a GRPO framework and employs policy shaping via regularized importance sampling. SRFT~\citep{fu2025srft} further unifies data mixing with a sample-level SFT loss in a single-stage framework that dynamically adjusts loss weights based on token entropy. CHORD~\citep{zhang2026onpolicy} reframes SFT as a dynamically weighted auxiliary objective within on-policy RL, employing a dual-control mechanism that combines a global coefficient for managing the transition from off-policy imitation to on-policy exploration with a token-wise weighting function.

In \textbf{expert-guided generation} methods, expert data is used to guide on-policy rollouts rather than mixed at the data level. UFT~\citep{liu2026uft} and BREAD~\citep{zhang2026bread} utilize SFT trajectories as prefixes for on-policy rollouts. UFT progressively masks the suffix of the expert demonstration, while BREAD initiates new rollouts by branching from intermediate steps. Prefix-RFT~\citep{huang2026blending} samples prefixes from demonstrations as guidance and employs an entropy-based clipping mechanism to select high-entropy expert tokens for SFT. TRAPO~\citep{su2026trustregion}  introduces Trust-Region SFT (TrSFT) to constrain policy updates within a trust region and employs an adaptive micro-group sampling mechanism that dynamically allocates expert guidance length based on return improvements.

In \textbf{interleaved SFT-RL update} methods, a third category interleaves RL updates with SFT steps on expert data. ReLIFT~\citep{ma2026learning} collects poorly performing samples during RL training and selectively applies SFT to challenging problems that RL fails to solve. HPT~\citep{lv2025hpt} dynamically decides whether to apply SFT or RL based on the model's rollout performance using binary weights. AMFT~\citep{he2025amft} views the balance between SFT and RL as a learnable parameter and introduces a meta-gradient adaptive weight controller.

%% file: chapter/method.tex
\section{Preliminaries}
We build our method upon the \textit{Mix-Policy GRPO} framework \citep{yan2026learning}, in which off-policy rollouts derived from expert data are directly injected into the group of on-policy rollouts generated by the current model. Thus, both on-policy and off-policy rollouts are simultaneously considered during advantage computation. Formally, the advantage $A_k$ for one rollout $\vy_k$ is computed as:
\newcommand{\on}{\text{on}}
\newcommand{\off}{\text{off}}
\newcommand{\old}{\text{old}}
\begin{equation} \label{eq:adv}
    A_k = \frac{R(\vy_k) - \mathrm{mean}(\gG_{\on}\cup\gG_{\off})}{\bcancel{\mathrm{std(\gG_{\on}\cup\gG_{\off})}}}, \quad
    \begin{cases}
        \gG_{\on} = \{R(\vy_i) | \vy_i \sim \pi_{\theta_{\old}}(\vx),i=1,...,N_{\on}\}, \\
        \gG_{\off} = \{R(\vy_j) | \vy_j \sim \pi_{\phi}(\vx),j=1,...,N_{\off}\}.
    \end{cases}
\end{equation}
where $R(\cdot)$ represents the reward function, $\vx$ is the input prompt, and $N_\on$ and $N_\off$ are the number of on-policy instances and off-policy instances in each group, respectively. 
In addition, $\pi_{\phi}$ denotes the off-policy expert distribution, $\pi_{\theta_\old}$ denotes the on-policy distribution where $\theta_\old$ is the model parameters before the update.
Following \citep{yan2026learning,su2026trustregion,ma2026learning}, we omit the standard deviation normalization term in the denominator in (\ref{eq:adv}).

Inheriting the standard GRPO objective \citep{shao2024deepseekmath}, the objective function on the updated model parameters $\theta$ is formulated as:
\begin{equation} \label{eq:luffy_loss}
\small
\begin{aligned}
    \gL_{\text{mix}}(\theta) = \frac{1}{Z}\left[\sum_{\vy_i \in \gY_\on}\sum_{t=1}^{|\vy_i|} \mathtt{CLIP}(r_{i,t},A_i, \epsilon)+\sum_{\vy_j \in \gY_\off}\sum_{t=1}^{|\vy_j|} \mathtt{CLIP}(\widehat{r}_{j,t},A_j, \epsilon)\right], 
    \begin{cases}
        r_{i,t} = \frac{\pi_{\theta}(y_{i,t}| \vx, \vy_{i,<t})}{\pi_{\theta_{\old}}(y_{i,t}| \vx, \vy_{i,<t})}, \\
        \widehat{r}_{j,t} = \frac{\pi_{\theta}(y_{j,t}| \vx, \vy_{j,<t})}{\pi_{\phi}(y_{j,t}| \vx, \vy_{j,<t})}.
    \end{cases}
\end{aligned}
\end{equation}
Here, $r_{i, t}$, $\widehat{r}_{j, t}$ are importance sampling ratios; $\gY_\on$, $\gY_\off$ denote the set of on-policy rollouts and off-policy rollouts in the group, respectively. In addition, $Z$ denotes the total number of tokens, and $\epsilon$ is the clipping coefficient. The clipped surrogate objective is given by $\mathtt{CLIP}(r, A, \epsilon) = \min(rA, \mathtt{clip}(r, 1-\epsilon,1+\epsilon)A)$. Given that the exact expert behavioral policy $\pi_\phi$ is typically inaccessible, we follow standard practice and set its probability value to $1$. 
One key issue of this framework is entropy collapse, as optimizing $\gL_{\text{mix}}(\theta)$ in (\ref{eq:luffy_loss}) leads to increasingly deterministic rollouts and a diminished capacity to explore diverse reasoning trajectories \citep{yan2026learning}.

\section{Methodology}
In this section, we introduce \textbf{MOER}, a mix-policy optimization framework that integrates decaying expert prefix guidance with a selective rollback mechanism (Sec. \ref{sec:expert_prefix}) and harmonized mix-policy objectives (Sec. \ref{sec:mix_obj}).

\subsection{Decaying Expert Prefix Guidance with Selective Rollback} \label{sec:expert_prefix}
\newcommand{\com}{\text{com}}
\newcommand{\pre}{\text{pre}}
To establish a better balance between autonomous self-exploration and expert guidance, we propose injecting an expert chain-of-thought (CoT) prefix directly into the input prompt, forcing the model to complete the remaining reasoning steps and final solutions. This completion process is formally defined as $\vy_{\com} \sim \pi_{\theta_{\old}}(\vx\oplus\vy^*_{\pre})$, where $\vy^*_{\pre}$ denotes the prefixed segment of the expert CoT. Ultimately, the concatenated sequence $\vy^*_{\pre}\oplus \vy_{\com}$ is treated as the off-policy rollout $\vy_{\off}$.

Furthermore, inspired by curriculum learning principles \citep{bengio2009curriculum}, we adopt a cosine annealing strategy to progressively shorten the injected expert prefix during training. This alleviates forgetting induced by static and full expert supervision and encourages the model to rely more heavily on its own self-exploration capabilities as training progresses. The prefix decaying schedule is defined as $\mathtt{CosAneal}(\vy) = \vy_{<r\cdot|\vy|}$, where $|\vy|$ is the number of tokens in $\vy$ and the prefix ratio $r$ is computed as
\newcommand{\warm}{\text{warm}}
\begin{equation} \label{eq:cos_aneal}\
r=\begin{cases}
        r_{\max}& \text{if}~ s<s_{\warm},\\
        r_{\min} + (r_{\max}-r_{\min})\left[0.5+0.5\cdot\cos\left(\frac{s-s_{\warm}}{S-s_{\warm}}\pi\right)\right] &\text{otherwise},\\
    \end{cases}
\end{equation}
where $r_{\min}$, $r_{\max}$, $s$, $s_{\warm}$, and $S$ denote the minimum prefix ratio, maximum prefix ratio, current training step, warm-up step, and maximum training steps, respectively. 

However, generating completions based on an expert prefix does not inherently guarantee the correctness of the final response. To mitigate this issue, we introduce a ``cost-free'' rollback mechanism. Specifically, if the completed response $\vy_\pre^* \oplus \vy_\com$ fails verification (i.e., the final solution is incorrect), we seamlessly roll back and replace the failed generation with the original full expert response $\vy^*$, which does not introduce computational overhead. 

\begin{figure}[t]
    \centering
    \vspace{-0.5em}
    \subfigure[Validation Accuracy]{
    \includegraphics[width=0.315\textwidth]{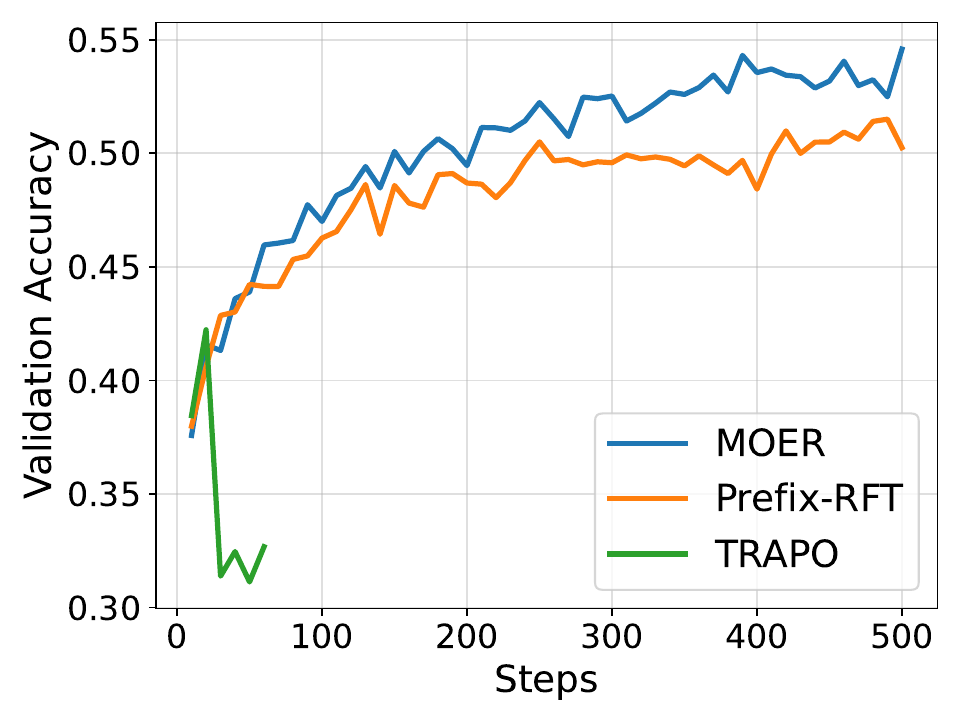}
    }
    \subfigure[Off-Policy Score]{
    \includegraphics[width=0.315\textwidth]{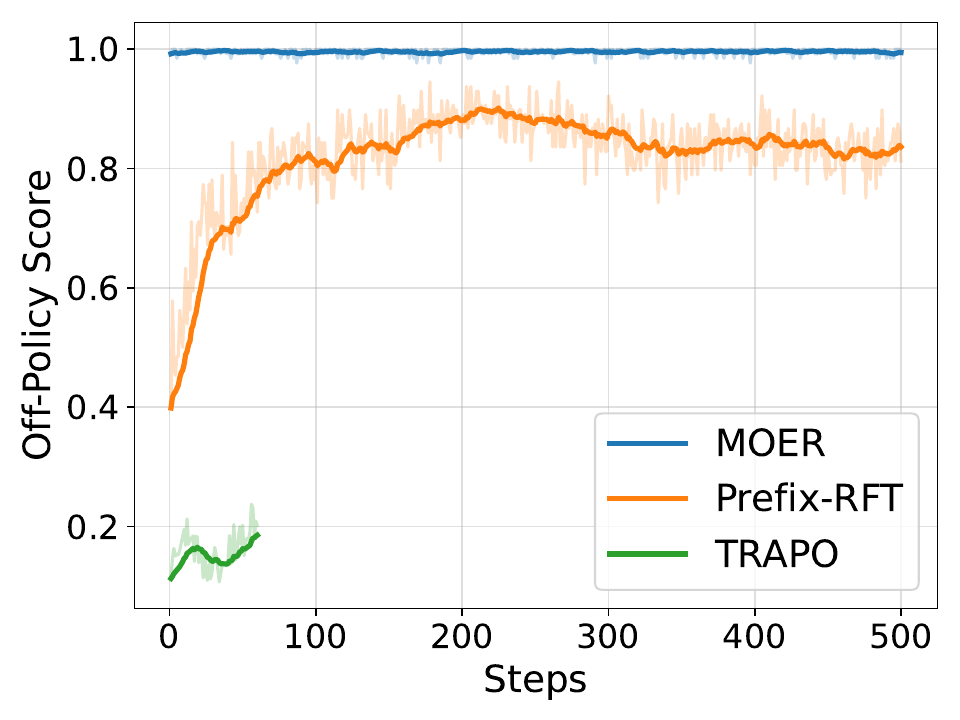}
    }
    \subfigure[Response Length]{
    \includegraphics[width=0.315\textwidth]{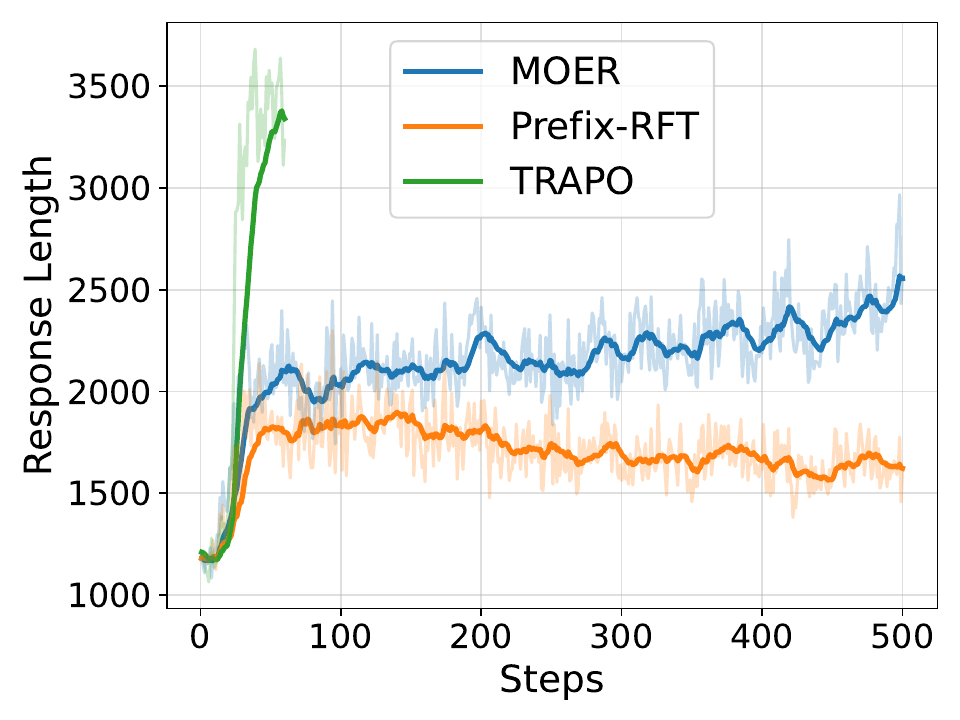}
    }
    \vspace{-0.5em}
    \caption{Training dynamics of MOER and other baselines with prefix guidance. The base model is Qwen2.5-Math-7B. \textbf{(a)} MOER reaches the highest validation accuracy. \textbf{(b)} Selective rollback mechanism guarantees the correctness of expert guidance. \textbf{(c)} MOER exhibits a stable increase in response length. Note that TRAPO is early-stopped due to training collapse.}
    \label{fig:analysis_prefix_baseline}
    \vspace{-1em}
\end{figure}

Compared to Prefix-RFT \citep{huang2026blending} and TRAPO \citep{su2026trustregion}, as illustrated in Figure \ref{fig:analysis_prefix_baseline}, our prefix-injection method with a selective rollback mechanism guarantees the correctness of the expert guidance and exhibits superior performance and stability. In contrast, Prefix-RFT achieves suboptimal performance and TRAPO collapses in the early stage of training with explosive increase in response length.

\subsection{Harmonized Mix-Policy Objective} \label{sec:mix_obj}
Recent studies \citep{yan2026learning,su2026trustregion,zhang2026onpolicy} have demonstrated that off-policy and on-policy tokens should be optimized via distinct loss formulations. Typically, existing methods modify the SFT loss function to dampen the gradient weights of off-policy tokens. Following \cite{su2026trustregion}, we incorporate the advantage term into a trust-region SFT formulation, where the gradient weight of low-probability tokens is clipped. The resulting off-policy loss function is defined as:
\begin{equation}\label{eq:off}
    \gL_{\text{off}}(\theta) = \sum_{\vy_j \in \gY_\off}\sum_{t=1}^{|\vy_j|} \frac{\pi_{\theta}(y_{j,t}| \vx, \vy_{j,<t})} {\max(\mathtt{sg}(\pi_{\theta}(y_{j,t}| \vx, \vy_{j,<t})), \gamma)}A_j,
\end{equation}
where the advantage $A_j$ is calculated by (\ref{eq:adv}), $\mathtt{sg}(\cdot)$ denotes the stop-gradient operator and $\gamma$ is the trust-region coefficient, set to $0.1$ by default. We defer the detailed discussion on different off-policy objectives proposed in prior work to Appendix \ref{app:discuss_off}. In the discussion, we suggest that all of these off-policy objectives can be seen as softened SFT objectives in essence. Furthermore, we conduct an ablation study on these objectives to corroborate this viewpoint.

For on-policy tokens, we leverage a modified GRPO loss featuring asymmetric clipping coefficients \citep{yu2026dapo} without standard deviation normalization in advantage computation \citep{liu2025understanding}, which is formulated as:
\newcommand{\low}{\text{low}}
\newcommand{\high}{\text{high}}
\begin{equation}
    \gL_{\text{on}}(\theta) = \sum_{\vy_i \in \gY_\on}\sum_{t=1}^{|\vy_i|}\min\left(r_{i,t}A_i,~ \mathtt{clip}(r_{i,t}, 1-\epsilon_{\low},1+\epsilon_{\high})A_i\right),
\end{equation}
where $\epsilon_{\low}$ and $\epsilon_{\high}$ are decoupled clipping coefficients. Notably, LUFFY does not adopt clipping operation for on-policy tokens, which leads to instability in the later stage of training. Therefore, we introduce the clipping operation back. Further analysis is deferred to Appendix \ref{app:analysis_clip}. Moreover, we empirically demonstrate the necessity of clipping in Figure \ref{fig:ab_hp} (c) of Appendix \ref{app:ab_hp}.

\begin{wrapfigure}{r}{0.42\textwidth}
    \centering
    \includegraphics[width=0.4\textwidth]{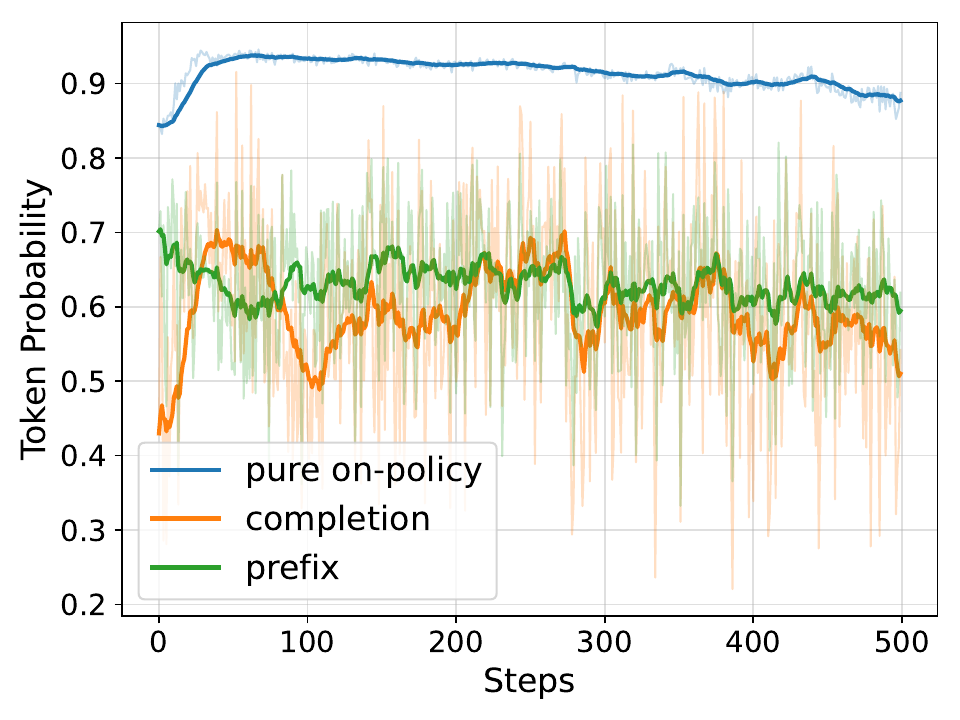}
    \vspace{-0.5em}
    \caption{Average output probability of pure on-policy, completion, and expert prefix tokens. The solid curves are smoothed.}
    \label{fig:token_prob}
    \vspace{-0.5em}
\end{wrapfigure}

In prior work \citep{su2026trustregion,huang2026blending}, the completion tokens are categorized as on-policy tokens. We argue that this categorization overlooks the distributional consequence of expert prefix conditioning. Although the completion $\vy_{\com}$ is sampled from $\pi_{\theta_{\old}}(\cdot|\vx\oplus\vy^*_{\pre})$ and is thus conditionally on-policy given the prefix, the prefix itself steers the model into context states that the current policy would rarely reach through its own exploration. Consequently, with respect to the output distribution induced by the original prompt $\vx$, the completion tokens are effectively off-policy. Furthermore, the selective rollback mechanism retains only verified completions, turning them into success-conditioned rollouts with an inherent selection bias. Empirically, as illustrated in Figure \ref{fig:token_prob}, the completion tokens exhibit substantially lower probabilities under the current policy than pure on-policy tokens, corroborating the output distribution shift. Optimizing such tokens with the standard on-policy objective introduces training instability. To address this issue, we treat $\vy_{\com}$ as off-policy tokens and compute their loss using the off-policy objective, which yields smoother gradient weights and serves as a stabilization strategy against the \textbf{expert-induced output distribution shift} and \textbf{selection bias}.

Ultimately, we combine $\gL_{\text{on}}$ and $\gL_{\text{off}}$ to formulate the harmonized mix-policy objective:
\begin{equation} \label{eq:loss_moer}
    \gL_{\text{MOER}}(\theta) =  \frac{1}{Z}(\gL_{\text{on}}(\theta) + \alpha\cdot\gL_{\text{off}}(\theta)),
\end{equation}
where $\alpha$ is the off-policy loss weight. Finally, the pseudocode of MOER is outlined in Algorithm \ref{alg:moer}.

\newcommand{\sol}{\text{sol}}
\begin{algorithm}[ht]
   \caption{\textbf{M}ix-Policy \textbf{O}ptimization under \textbf{E}xpert Prefix Guidance with \textbf{R}ollback (\textbf{MOER})}
   \label{alg:moer}
   \setstretch{1.05}
\begin{algorithmic}[1]
    \STATE {\bfseries Input:} parameter $\theta$, dataset $\gD$, number of on-policy rollouts $N_{\on}$, number of off-policy rollouts $N_{\off}$, clipping coefficients $(\epsilon_{\low}, \epsilon_{\high})$, trust-region coefficient $\gamma$, off-policy loss weight $\alpha$.
    
    \FOR{each mini-batch $\gB \in \gD$}
    \FOR{each prompt and expert answer $(\vx,\vy^*) \in\gB$} 
    
    \STATE $\vy^*_{\cot}, \vy^*_{\sol} \leftarrow \mathtt{Split}(\vy^*)$\COMMENT{Split the answer into chain-of-thought and solution}
    \STATE $\vy^*_{\pre} \leftarrow \mathtt{CosAneal}(\vy^*_{\cot})$ via Eq. (\ref{eq:cos_aneal})
    \FOR{$i=1$ to $N_{\on}$}  
        \STATE $\vy_{\on}^i \sim \pi_{\theta_{\old}}(\cdot|\vx)$\COMMENT{On-policy rollout}
    \ENDFOR
    \FOR{$j=1$ to $N_{\off}$}  
        \STATE $\vy_{\com}^j \sim \pi_{\theta_{\old}}(\cdot|\vx\oplus\vy^*_{\pre})$ \COMMENT{Completing the response based on expert prefix}
        \STATE $ \vy_{\off}^j\leftarrow \vy^*_{\pre} \oplus \vy_{\com}^j$~~\textbf{if}~~$\vy^*_{\pre}\oplus \vy_{\com}^j$ is correct~~\textbf{else}~~$\vy^*$ \COMMENT{Selective rollback}
    \ENDFOR
    \STATE Calculate advantages $\{A_i\}_{i=1}^{N_{\on}+N_{\off}}$ via Eq. (\ref{eq:adv})
    \ENDFOR
    \STATE Calculate $\gL_{\text{MOER}}$ via Eq. (\ref{eq:loss_moer})
    \STATE Update $\theta$ based on $\gL_{\text{MOER}}$
    \ENDFOR
\end{algorithmic}
\end{algorithm}

%% file: chapter/exp.tex
\section{Experiments}
\subsection{Experimental Settings} \label{sec:exp_setting}
\textbf{Training Details.} We focus on mathematical reasoning task in this section and defer the experiments on coding task to Appendix \ref{app:code}. The training set is OpenR1-Math-46k-8192 \citep{yan2026learning}, where the prompts are collected from NuminaMath 1.5 \citep{numina_math_datasets}, and the expert traces are generated by DeepSeek-R1. Following prior work \citep{yan2026learning, su2026trustregion, huang2026blending}, Qwen2.5-Math-7B \citep{yang2024qwen2} is adopted as the base model. We also evaluate our method on Qwen2.5-Math-1.5B, Qwen2.5-Instruct-7B, and Qwen2.5-Instruct-14B \citep{qwen2025qwen25technicalreport}. 

\textbf{Implementation Details.} For on-policy loss, we use a modified GRPO loss featuring decoupled clipping coefficients \citep{yu2026dapo} without standard deviation normalization in advantage computation \citep{liu2025understanding}. The clipping coefficients $\epsilon_{\low}$ and $\epsilon_{\high}$ are set to $0.2$ and $0.24$, respectively. The KL regularization term is removed by setting $\beta=0$ and the entropy loss coefficient is set to $0.01$. For off-policy loss, the trust-region coefficient $\gamma$ is set to $0.1$. The off-policy loss weight $\alpha$ is set to $0.2$ for Qwen2.5-Math-7B and Qwen2.5-Math-1.5B, and is set to $0.1$ for Qwen2.5-Instruct-7B. For the prefix decaying schedule, the warm-up step $s_{\warm}=50$, the minimum prefix ratio $r_{\min}=0.1$, the maximum prefix ratio $r_{\max}=1.0$. The rollout batch size is $128$ and the update batch size is $64$. Other unspecified hyperparameters are consistent with those in LUFFY \citep{yan2026learning}.

\textbf{Evaluation.} We evaluate the models with the highest validation accuracy on both math and general domain reasoning benchmarks. Specifically, the math benchmarks include AIME 2024, AIME 2025, AMC \citep{numina_math_datasets}, MATH-500 \citep{hendrycks2021measuring}, Minerva \citep{lewkowycz2022solving}, and OlympiadBench \citep{He2024OlympiadBenchAC}. For AIME 2024, AIME 2025 and AMC, we report avg@32 as the test set is relatively small, and for the other three benchmarks, we report pass@1. For general domain reasoning benchmarks, we include the pass@1 on ARC-c \citep{allenai:arc}, GPQA-diamond \citep{rein2024gpqa}, and MMLU-Pro \citep{wang2024mmlupro}.

\textbf{Baselines.} We comprehensively compare MOER with various baselines, including \textbf{(1)} SFT: train the model with the same dataset using SFT; \textbf{(2)} RL: RL training using standard GRPO loss \citep{shao2024deepseekmath}; \textbf{(3)} RL w/ SFT Loss: combining SFT loss with GRPO loss during RL training; \textbf{(4)} SFT-then-RL: two-stage training that continues RL training with GRPO loss after SFT. Furthermore, we include other mix-policy baselines, e.g., \textbf{(5)} LUFFY \citep{yan2026learning}: insert a full expert trace into on-policy rollout group; \textbf{(6)} ReLIFT \citep{ma2026learning}: alternating training between RL and SFT; \textbf{(7)} CHORD \citep{zhang2026onpolicy}: GRPO loss + reweighted SFT loss; \textbf{(8)} TRAPO \citep{su2026trustregion}: dynamically allocate expert prefix length based on return improvements; \textbf{(9)} Prefix-RFT \citep{huang2026blending}: randomly sample expert prefix and SFT update on off-policy tokens with high entropy.

\subsection{Main Results}
\begin{table*}[t]
\centering
\setlength{\tabcolsep}{3.5pt}
\vspace{-1em}
\caption{Main experiment results on math and general reasoning benchmarks based on Qwen2.5-Math-7B. All baselines are evaluated on the same dataset and base model. \textbf{Bold} and \ul{underline} indicate the best and second-best results, respectively.}
\label{tab:main_results}
\vspace{0.5em}
\resizebox*{\textwidth}{!}{
\begin{tabular}{l ccccc >{\columncolor{yellowavg}}c | ccc  >{\columncolor{blueavg}}c}
\toprule
\multirow{2}{*}{\textbf{Method}} & \multicolumn{6}{c}{\textbf{Math Reasoning Performance}} & \multicolumn{4}{c}{\textbf{General Domain Reasoning Performance}} \\
\cmidrule(lr){2-7} \cmidrule(lr){8-11}
 & \textbf{AIME 24/25} & \textbf{AMC} & \textbf{MATH-500} & \textbf{Minerva} & \textbf{Olympiad} & \textbf{Avg.} & \textbf{ARC-c} & \textbf{GPQA\textsuperscript{*}} & \textbf{MMLU-Pro} & \textbf{Avg.} \\
\midrule
Base & 11.5/4.9 & 31.3 & 43.6 & 7.4 & 15.6 & 19.0 & 18.2 & 11.1 & 16.9 & 15.4 \\
\midrule
\multicolumn{11}{c}{\textbf{Basic Baselines}} \\
\midrule
SFT & 22.2/\ul{22.3} & 52.8 & 82.6 & \textbf{40.8} & 43.7 & 44.1 & 75.2 & 24.7 & 42.7 & 47.5 \\
RL & 25.1/15.3 & 62.0 & 84.4 &39.3 & 46.8 & 45.5 & 82.3 & \textbf{40.4} & 49.3 & 57.3 \\
RL w/ SFT Loss & 19.5/16.4 & 49.7 & 80.4 & 34.9 & 39.4 & 40.1 & 71.2 & 23.7 & 43.2 & 46.0 \\
SFT-then-RL & 25.8/\textbf{23.1} & 62.7& 87.2 & 39.7 & 50.4 & 48.2 & 72.4 & 24.2 & 37.7 & 44.8 \\
\midrule
\multicolumn{11}{c}{\textbf{Mix-policy Baselines}} \\
\midrule
LUFFY & \textbf{28.5}/21.6 & 61.0 & 86.4 & \ul{40.4} & 52.4 & 48.4 & 81.9 & 37.9 & \ul{52.2} & 57.3 \\
ReLIFT & 26.5/20.2 & \ul{63.5} & \textbf{88.0} & 39.3 & \ul{53.9} & \ul{48.6} & 82.3 & 37.4 & 52.1 & 57.3\\
CHORD & 25.1/16.6 & 59.0 & 83.4 & 35.7 & 51.1 & 45.2 & \ul{82.4} & 38.4 & 51.4 & \ul{57.4}\\
TRAPO & 0.2/0.3 & 1.9 & 5.4 & 2.9 & 1.2 & 2.0 & 0.1 & 0.0 & 3.0 & 1.0\\
Prefix-RFT & 19.2/17.1 & 58.9 & 83.8 & 36.0 & 50.8 & 44.3 & 81.5 & 35.9 & 50.4 & 55.9\\
\midrule
\multicolumn{11}{c}{\textbf{Our Method}} \\
\midrule
MOER & \ul{27.9}/22.2& \textbf{63.7}& \ul{86.8}& 38.6& \textbf{55.6}& \textbf{49.1}& \textbf{84.0}& \ul{38.9}& \textbf{54.1}& \textbf{59.0}\\
\bottomrule
\end{tabular}
}
\vspace{-1em}
\end{table*}

We first evaluate MOER against basic and mix-policy baselines using the Qwen2.5-Math-7B base model. To ensure a fair comparison, all mix-policy baselines are reproduced on the same dataset and base model as our method.
As reported in Table~\ref{tab:main_results}, LUFFY and ReLIFT are the only mix-policy baselines that marginally outperform SFT-then-RL. In contrast, MOER achieves an average score of \textbf{49.1} across six mathematical reasoning benchmarks, outperforming all baseline models. This performance gap demonstrates that MOER effectively balances the internalization of expert knowledge with self-exploration, resulting in a better acquisition of reasoning capabilities.

Furthermore, MOER exhibits superior generalization on out-of-distribution (OOD) general-domain reasoning benchmarks, achieving an average score of \textbf{59.0}. While the SFT-then-RL baseline achieves high in-domain performance, its unsatisfactory performance on OOD tasks indicates overfitting during the initial SFT stage. These results confirm that MOER successfully harmonizes the strengths of both paradigms, acquiring specialized skills without sacrificing general task utility.

Finally, alternative prefix-based methods such as TRAPO and Prefix-RFT suffer from training collapse or suboptimal performance. The instability of TRAPO stems from overly aggressive off-policy supervision, where up to 4 out of 8 group rollouts consist of off-policy traces in the worst case. Meanwhile, Prefix-RFT yields suboptimal results due to the lack of correctness guarantees for model completions, which ultimately causes it to degenerate into GRPO.

\subsection{Results on More Models}
We further compare our method with the most competitive baselines, LUFFY and ReLIFT, across different model scales and base model categories (math-specialized vs. general instruction-tuned). Note that the results on Qwen2.5-Instruct-14B are deferred to Appendix \ref{app:14b}. As shown in Table~\ref{tab:more_models} and Table \ref{tab:14b} in Appendix \ref{app:14b}, MOER consistently demonstrates superior or competitive performance across both model scales and base model categories.

On Qwen2.5-Math-1.5B, MOER ties with LUFFY for the highest average math score of \textbf{38.1}, while outperforming ReLIFT (34.1). This plateau in math performance relative to LUFFY can likely be attributed to the constrained parameter capacity of the 1.5B model, which restricts the potential gain from advanced self-directed exploration in highly complex mathematical reasoning. Nevertheless, MOER retains significantly stronger general domain capabilities, achieving a domain-leading average score of \textbf{40.6}, indicating its competitive performance on OOD tasks.

On Qwen2.5-Instruct-7B, the advantages of MOER become even more pronounced. In mathematical reasoning, MOER attains the highest average score of \textbf{42.3}. 
More importantly, MOER exhibits exceptional stability and general utility preservation on general domain reasoning tasks, achieving an overall average of \textbf{63.8}. Notably, while LUFFY suffers a severe performance collapse on ARC-c and GPQA, MOER maintains robust performance across all benchmarks. 

\begin{table*}[t]
\centering
\setlength{\tabcolsep}{3.5pt}
\vspace{-1em}
\caption{Experiment results on math and general reasoning benchmarks based on Qwen2.5-Math-1.5B and Qwen2.5-Instruct-7B.}
\label{tab:more_models}
\vspace{0.5em}
\resizebox*{\textwidth}{!}{
\begin{tabular}{l ccccc >{\columncolor{yellowavg}}c | ccc  >{\columncolor{blueavg}}c}
\toprule
\multirow{2}{*}{\textbf{Method}} & \multicolumn{6}{c}{\textbf{Math Reasoning Performance}} & \multicolumn{4}{c}{\textbf{General Domain Reasoning Performance}} \\
\cmidrule(lr){2-7} \cmidrule(lr){8-11}
 & \textbf{AIME 24/25} & \textbf{AMC} & \textbf{MATH-500} & \textbf{Minerva} & \textbf{Olympiad} & \textbf{Avg.} & \textbf{ARC-c} & \textbf{GPQA\textsuperscript{*}} & \textbf{MMLU-Pro} & \textbf{Avg.} \\
\midrule
\multicolumn{11}{c}{\textit{Qwen2.5-Math-1.5B}} \\
\midrule
LUFFY & 16.0/\textbf{12.9}& \textbf{47.1}& 77.2& \textbf{32.7}& 42.7& \textbf{38.1}& 55.6& \textbf{26.8}& \textbf{34.6}& 39.0 \\
ReLIFT & 14.6/9.0& 41.7& 72.0& 29.0& 38.4& 34.1& 55.0& 25.8& 30.4& 37.1\\
\textbf{MOER} &\textbf{16.6}/12.2& 46.7& \textbf{79.6}& 29.8& \textbf{43.4}& \textbf{38.1}& \textbf{60.4}& \textbf{26.8}& \textbf{34.6}& \textbf{40.6}\\
\midrule
\multicolumn{11}{c}{\textit{Qwen2.5-Instruct-7B}} \\
\midrule
LUFFY & \textbf{18.3}/15.2& 50.2& \textbf{81.6}& 37.1& 46.5& 41.5& 2.7& 1.0& \textbf{60.7}& 21.5 \\
ReLIFT & 18.0/\textbf{16.1}&51.1&81.0&37.5 &44.9&41.4&\textbf{89.9}&32.3&56.6&59.6\\
\textbf{MOER} &15.9/15.6 &\textbf{51.7}& 81.4&\textbf{39.0}&\textbf{50.1}&\textbf{42.3}&89.1&\textbf{43.4}& 59.0&\textbf{63.8}\\
\bottomrule
\end{tabular}
}
\vspace{-0.5em}
\end{table*}

\subsection{Ablation Study and Analysis} \label{sec:ab}
We conduct ablation study and detailed analysis on the core components of our method in this section. In addition, we study how different hyperparameters and computation budgets affect performance in Appendix~\ref{app:ab_hp} and \ref{app:pass}, respectively.

To numerically investigate the contributions of individual components in MOER, we construct a series of ablation variants starting from the baseline LUFFY model and incrementally integrating our proposed modules based on Qwen2.5-Math-7B. 
As reported in Table~\ref{tab:ab_component}, starting from the base LUFFY model (48.4 math average, 57.3 general average), applying the harmonized objective function (+\textit{HO}) yields consistent improvements on both math and general domain performance (48.6 and 58.3). Adding prefix decaying (+\textit{HO}+\textit{PG}+\textit{COff}) leads to a decline in math performance to 46.2 due to the lack of correctness guarantees for model
completions. When rollback is further incorporated (+\textit{HO}+\textit{PG}+\textit{SR}), the mathematical performance recovers substantially to 48.9 (see Appendix \ref{app:motivation} for a detailed discussion of the rollback dynamics along training).
Additionally, removing expert guidance but with rollback (+\textit{HO}-\textit{FG}+\textit{SR}), i.e., replacing one incorrect on-policy rollout with full expert trace, results in a noticeable drop in performance.
Finally, our complete framework (+\textit{HO}+\textit{PG}+\textit{SR}+\textit{COff}), which treats completion tokens as off-policy tokens, achieves the highest overall performance on both math and general domain reasoning (\textbf{49.1} and \textbf{59.0}). The detailed analysis on the effect of each core component is presented as follows:

\begin{table*}[t]
\centering
\small
\setlength{\tabcolsep}{3.5pt}
\vspace{-0.5em}
\caption{Ablation study on components of MOER based on Qwen2.5-Math-7B. \textit{HO} denotes \textbf{harmonized objective}, \textit{PG} denotes \textbf{decaying prefix guidance}, \textit{SR} denotes \textbf{selective rollback}, \textit{FG} denotes \textbf{full expert guidance}, \textit{COff} denotes \textbf{completion as off-policy tokens}.}
\label{tab:ab_component}
\vspace{0.5em}
\resizebox*{\textwidth}{!}{
\begin{tabular}{l ccccc >{\columncolor{yellowavg}}c | ccc  >{\columncolor{blueavg}}c}
\toprule
\multirow{2}{*}{\textbf{Method}} & \multicolumn{6}{c}{\textbf{Math Reasoning Performance}} & \multicolumn{4}{c}{\textbf{General Domain Reasoning Performance}} \\
\cmidrule(lr){2-7} \cmidrule(lr){8-11}
 & \textbf{AIME 24/25} & \textbf{AMC} & \textbf{MATH-500} & \textbf{Minerva} & \textbf{Olympiad} & \textbf{Avg.} & \textbf{ARC-c} & \textbf{GPQA\textsuperscript{*}} & \textbf{MMLU-Pro} & \textbf{Avg.} \\
\midrule
LUFFY (GRPO+\textit{FG}) & 28.5/21.6 & 61.0 & 86.4 & 40.4 & 52.4 & 48.4 & 81.9 & 37.9 & 52.2 & 57.3 \\
~+\textit{HO} & 27.8/21.5& 63.1& 87.6& 37.9& 53.8& 48.6&83.2 &38.9 & 52.9& 58.3\\
~+\textit{HO}+\textit{PG}+\textit{COff} & 24.7/15.4 & 60.7 & 84.4 & 40.8 & 51.3 & 46.2 & 82.5 & 41.9 & 50.4 & 58.3\\
~+\textit{HO}+\textit{PG}+\textit{SR} & 27.1/22.6& 63.4	& 87.8& 38.2& 54.4& 48.9& 83.3& 39.5& 53.2& 58.7\\
~+\textit{HO}-\textit{FG}+\textit{SR} & 24.9/18.8 & 61.9 & 85.8 & 37.5 & 53.9 & 47.1 & 83.4 & 36.4 & 53.3 & 57.7\\
~+\textit{HO}+\textit{PG}+\textit{SR}+\textit{COff} (\textbf{MOER})& 27.9/22.2& 63.7& 86.8& 38.6& 55.6& \textbf{49.1}& 84.0& 38.9& 54.1& \textbf{59.0}\\
\bottomrule
\end{tabular}
}
\vspace{-1em}
\end{table*}

\begin{figure}[t]
\vspace{-0.5em}
    \centering
    \subfigure[Validation Accuracy]{
    \includegraphics[width=0.315\textwidth]{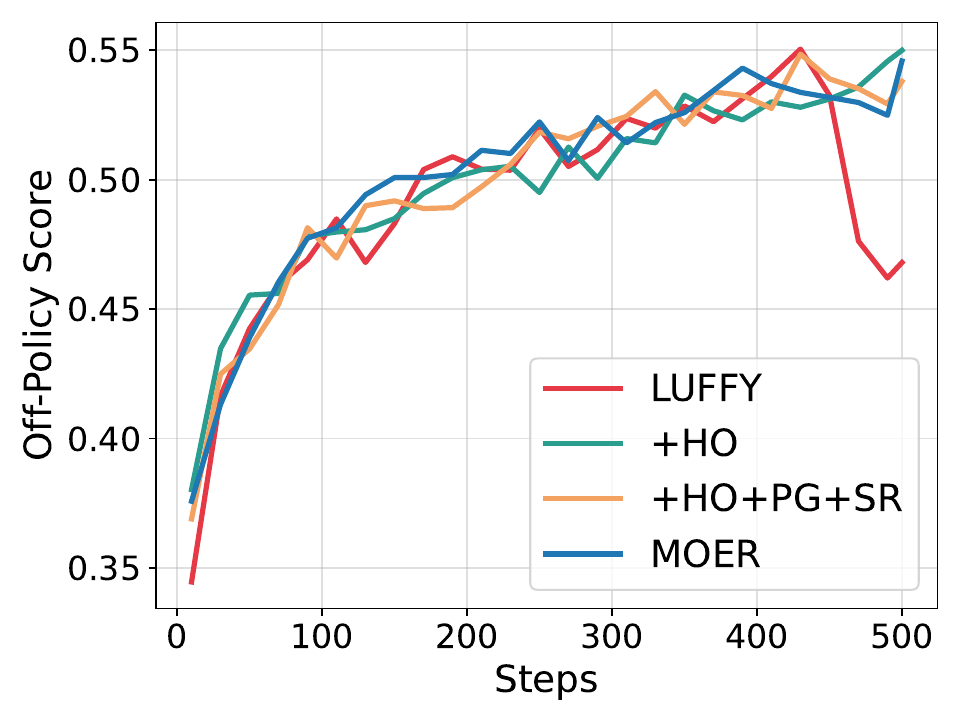}
    }
    \subfigure[Entropy]{
    \includegraphics[width=0.315\textwidth]{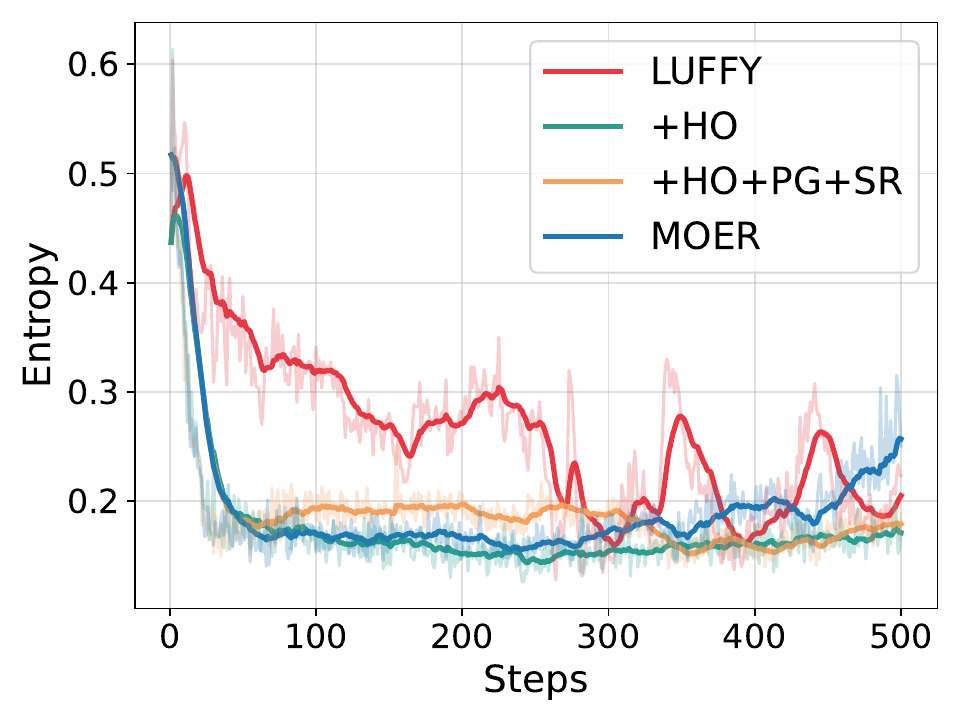}
    }
    \subfigure[Response Length]{
    \includegraphics[width=0.315\textwidth]{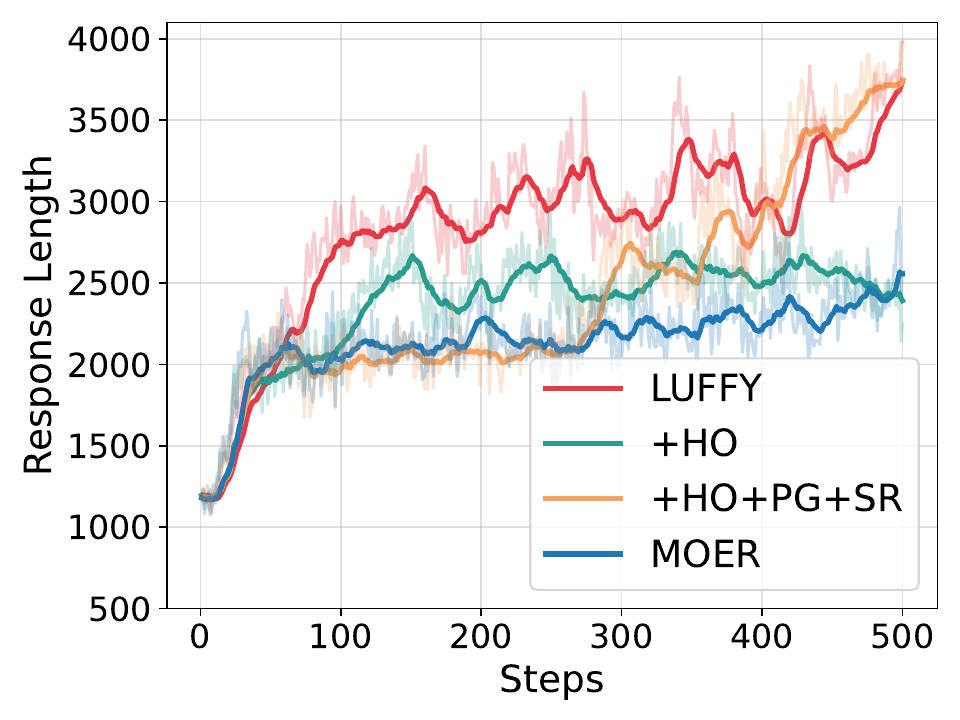}
    }
    \vspace{-0.5em}
    \caption{Training dynamics of MOER and its ablations. \textit{HO} denotes harmonized objective, \textit{PG} denotes decaying prefix guidance, and \textit{SR} denotes selective rollback.}
    \label{fig:analysis_ab}
\end{figure}

\textbf{Effect of Harmonized Objective (HO).}
As illustrated in Figure~\ref{fig:analysis_ab} (a)-(c), our harmonized objective effectively mitigates fluctuations of entropy and response length, preventing late-stage training collapse and ultimately yielding superior reasoning performance. A key factor driving this stability is the asymmetric clipping mechanism applied to the importance sampling ratio. As analyzed in Appendix~\ref{app:analysis_clip}, LUFFY's unclipped design introduces training instability that intensifies when updates become more ``off-policy". Conversely, the symmetric clipping in standard GRPO induces entropy collapse. Furthermore, our extended discussion in Appendix~\ref{app:discuss_off} reveals that various off-policy objectives proposed in prior work share a similar stabilizing effect during training, as they essentially act as softened SFT objectives.

\textbf{Effect of Decaying Prefix Guidance (PG).}
Decaying prefix guidance is designed to alleviate forgetting induced by static and full expert supervision and progressively foster self-exploration as training proceeds.
The results presented in Table \ref{tab:ab_component} indicates that MOER exhibits better OOD performance than the variant utilizing full expert guidance (\textit{+HO}).
As shown by the entropy curve in Figure~\ref{fig:analysis_ab} (b), MOER maintains strong exploratory behavior in the later stage of training. In contrast, \textit{+HO} exhibits a weaker response diversity, which is evidenced by a smaller entropy.
Additionally, as discussed in Appendix~\ref{app:motivation}, the diminishing accuracy gap between on-policy rollouts and completions furhter supports the design of the decaying prefix schedule.

\begin{figure}[t]
\vspace{-0.5em}
    \centering
    \subfigure[Entropy]{
    \includegraphics[width=0.315\textwidth]{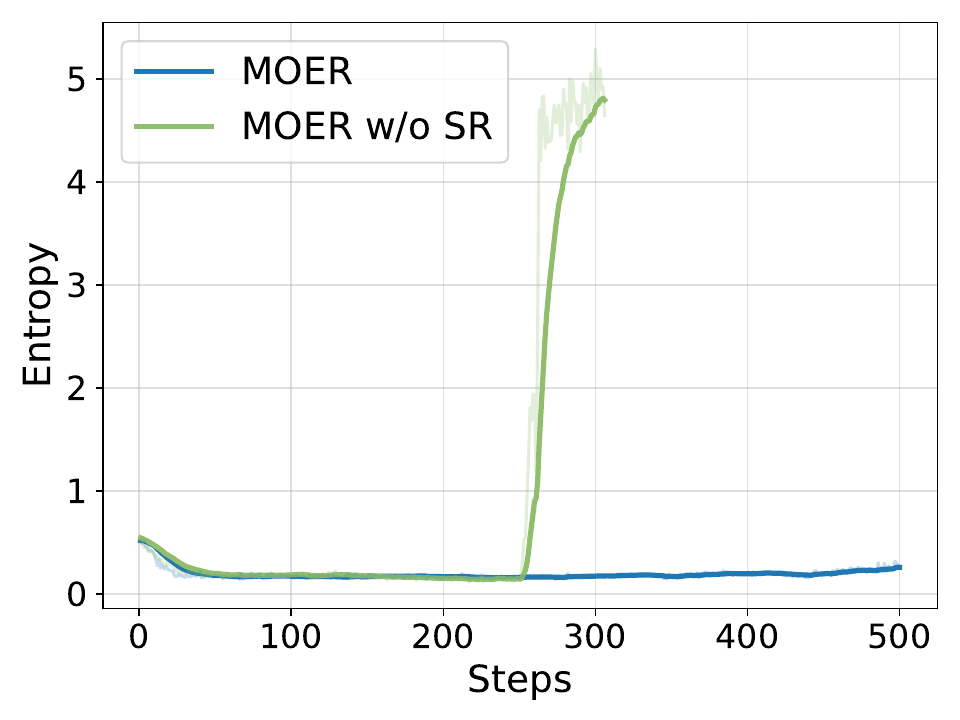}
    }
    \subfigure[Off-Policy Score]{
    \includegraphics[width=0.315\textwidth]{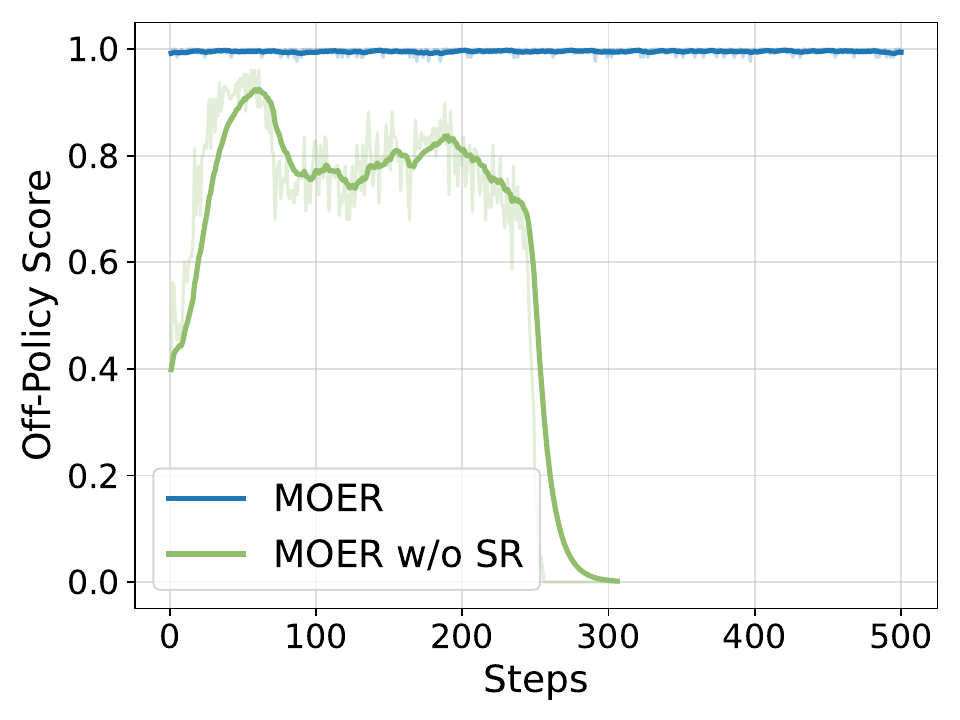}
    }
    \subfigure[Actual Prefix Ratio]{
    \includegraphics[width=0.315\textwidth]{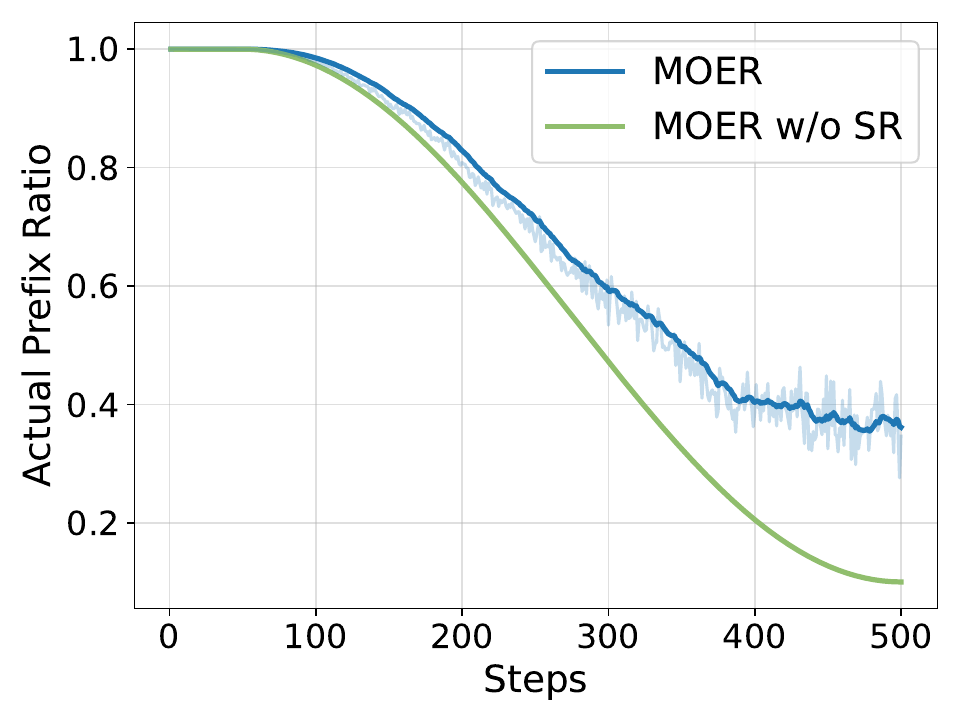}
    }
    \vspace{-0.5em}
    \caption{Effect of selective rollback mechanism. Note that MOER w/o \textit{SR} is identical to \textit{+HO+PG+COff} and it is early-stopped due to training collapse. The actual prefix ratio is calculated by $\mathrm{prefix~ratio} \times (1-\mathrm{rollback~rate}) + \mathrm{rollback~rate}$. 
    }
    \label{fig:analysis_sr}
    \vspace{-0.5em}
\end{figure}

\textbf{Effect of Selective Rollback (SR).}
As depicted in Figure~\ref{fig:analysis_sr} (a) and (b), removing the selective rollback mechanism while retaining decaying prefix guidance (\textit{+HO+PG+COff}) leads to training collapse, characterized by an off-policy score dropping to zero alongside explosive entropy growth. This finding demonstrates that the efficacy of decaying prefix guidance relies inherently on the rollback mechanism. Moreover, Figure~\ref{fig:analysis_sr} (c) confirms that the rollback mechanism does not degenerate into static expert guidance, as the actual prefix ratio steadily decays over the course of training.

\textbf{Effect of Completions as Off-Policy Tokens (COff).}
Treating completion tokens $\vy_{\text{com}}$ purely as on-policy tokens, i.e., computing their updates via standard on-policy loss functions (\textit{+HO+PG+SR}), introduces late-stage training instability with excessively increasing response length (see Figure~\ref{fig:analysis_ab} (c)). This instability ultimately leads to suboptimal final performance and overlong response length, underscoring the necessity of treating completion tokens as off-policy supervision.

%% file: chapter/conclusion.tex
\section{Conclusion}
In this paper, we propose MOER, a mix-policy optimization framework that harmonizes expert guidance with self-exploration for LLM reasoning. MOER combines decaying expert prefix guidance with selective rollback mechanism that guarantees the correctness of off-policy trajectories, and a harmonized mix-policy objective that optimizes on- and off-policy tokens with distinct loss formulations. Experiments across model scales and categories demonstrate that MOER achieves state-of-the-art math reasoning performance while preserving strong general-domain capabilities.


%% file: appendix/analysis.tex
\section{Discussion on Different Off-Policy Objectives} \label{app:discuss_off}
\begin{table}[h]
\centering
\small
\vspace{-1em}
\caption{Properties of gradient weights of different off-policy objectives. \textit{Scaled} denotes that the gradient weight is scaled based on the SFT objective. \textit{Bounded} denotes that the value of the gradient weight is bounded.}
\vspace{0.5em}
\begin{tabular}{c|c c c c c c}
    \toprule
     & LUFFY & ReLIFT & CHORD & TRAPO & Prefix-RFT & MOER\\
    \midrule
    \textit{Scaled} & \cmark & \xmark & \cmark & \xmark & \cmark & \cmark\\
    \textit{Bounded} & \cmark & \xmark & \cmark & \cmark & \xmark & \cmark\\
    \bottomrule
\end{tabular}
\label{tab:off_policy}
\end{table}

We first summarize the properties of the off-policy objectives adopted across various mix-policy baselines in Table~\ref{tab:off_policy}, with their corresponding gradient weight curves illustrated in Figure~\ref{fig:gradient_weight}. A detailed analysis of these objectives is presented below:

\textbf{LUFFY} \citep{yan2026learning}.
Omitting the advantage term $A$, the single-token off-policy objective in LUFFY is defined as $\gL_{\text{LUFFY}} = -\pi_{\theta} / (\pi_{\theta} + \gamma)$, where $\gamma = 0.1$ and $\pi_{\theta}$ denotes the token probability. The gradient with respect to parameter $\theta$ is $\nabla_{\theta}\gL_{\text{LUFFY}} = -\frac{\gamma}{(\pi_{\theta}+\gamma)^2} \nabla_{\theta}\pi_{\theta}$, yielding a gradient weight of $w_{\text{LUFFY}} = \gamma / (\pi_{\theta}+\gamma)^2$. Since $\pi_\theta \in [0, 1]$, the weight range is $w_{\text{LUFFY}} \in [\gamma / (1+\gamma)^2, 1/\gamma]$, making it both scaled and bounded.

\textbf{ReLIFT} \citep{ma2026learning}.
The off-policy objective in ReLIFT is formulated as $\gL_{\text{ReLIFT}} = \gL_{\text{CE}} + \alpha \gL_{\text{Entropy}} = -(1 + \alpha\pi_{\theta}) \log\pi_{\theta}$, where $\alpha = 1\times10^{-4}$. The corresponding gradient weight is $w_{\text{ReLIFT}} = 1/\pi_{\theta} + \alpha(\log\pi_{\theta} + 1) \in [1 + 2\alpha, \infty)$. Given the minuscule magnitude of $\alpha$, this gradient weight curve closely tracks that of the standard SFT objective. Consequently, it is neither scaled nor bounded.

\textbf{CHORD} \citep{zhang2026onpolicy}.
The off-policy objective in CHORD is defined as $\gL_{\text{CHORD}} = -\mathtt{sg}(\pi_\theta(1-\pi_\theta)) \log\pi_{\theta}$, where $\mathtt{sg}(\cdot)$ denotes the stop-gradient operator. The resulting gradient weight is $w_{\text{CHORD}} = 1 - \pi_\theta \in [0, 1]$, which is both scaled and bounded.

\textbf{TRAPO} \citep{su2026trustregion}.
TRAPO introduces a trust-region SFT objective identical to Eq.~(\ref{eq:off}), but it lacks scaling via a coefficient $\alpha$ as in our approach. Thus, its gradient weight is $w_{\text{TRAPO}} = 1 / \max(\pi_\theta, \gamma) \in [1, 1/\gamma]$, where $\gamma = 0.1$. This weight is bounded but unscaled.

\textbf{Prefix-RFT} \citep{huang2026blending}.
The off-policy objective in Prefix-RFT is given by $\gL_{\text{Prefix-RFT}} = \pi_{\theta} / \mathtt{sg}(\pi_\theta)$, applied selectively to the top-$20\%$ highest-entropy tokens. As a result, the expected gradient weight satisfies $\mathbb{E}[w_{\text{Prefix-RFT}}] \approx 0.2 / \pi_\theta$, which is scaled but unbounded.

As illustrated in Figure~\ref{fig:gradient_weight}, the gradient weight curves of all scaled objectives nearly overlap in the high token-probability regime ($\pi_\theta \to 1$). Divergence primarily occurs in low-probability regions, where LUFFY and Prefix-RFT assign significantly larger gradient weights than CHORD and MOER. Despite these structural variations, all considered off-policy objectives essentially function as softened SFT objectives.

\begin{figure}[b]
\vspace{-0.5em}
    \centering
    \begin{minipage}[c]{0.42\textwidth}
        \centering
        \vspace{0.5em}
        \includegraphics[width=\linewidth]{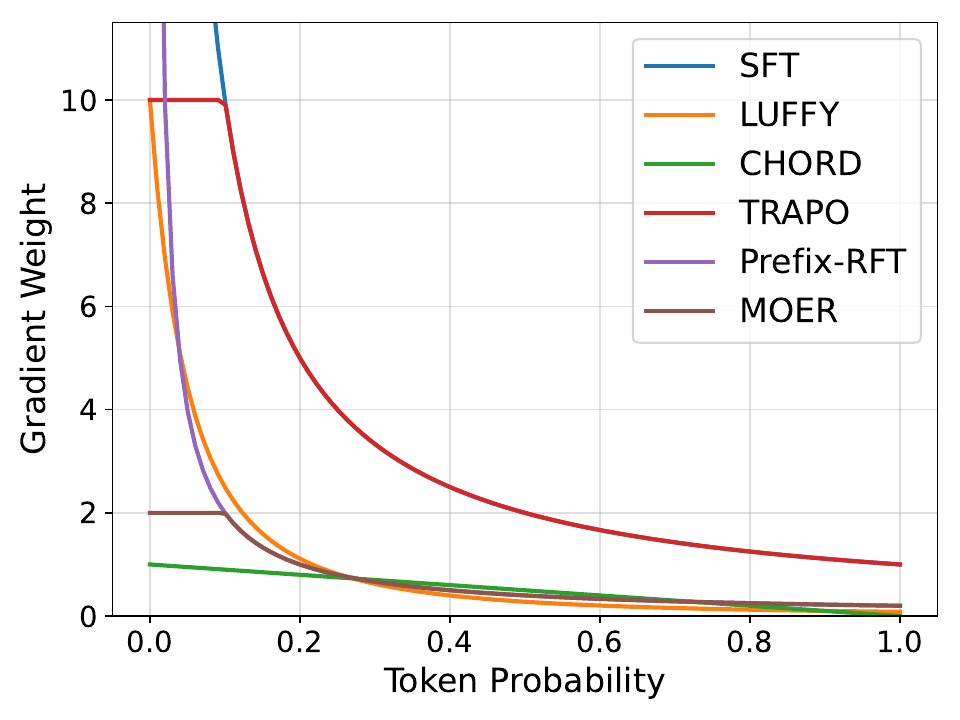}
        \vspace{-2em}
        \caption{Gradient weight curves of different off-policy objectives.}
        \label{fig:gradient_weight}
    \end{minipage}
    \hfill
    \begin{minipage}[c]{0.5\textwidth}
    \vspace{-1em}
        \centering
        \small
        \setlength{\tabcolsep}{4pt}
        \captionof{table}{Performance of MOER with scaled off-policy objectives on math and general domain reasoning benchmarks (average scores).}
        \label{tab:off_obj_ab}
        \vspace{0.8em}
        \begin{tabular}{l >{\columncolor{yellowavg}}c >{\columncolor{blueavg}}c}
            \toprule
            \textbf{Off-Policy Objective} & \textbf{Math} & \textbf{General}\\
            \midrule
            \textbf{MOER} & \textbf{49.1} & \textbf{59.0}\\
            \midrule
            ~w/ \textit{LUFFY} Loss & 48.2&58.0 \\
            ~w/ \textit{CHORD} Loss & 48.9  & 56.5\\
            ~w/ \textit{Prefix-RFT} Loss &46.9 & 57.1\\
            \bottomrule
        \end{tabular}
    \end{minipage}
    \vspace{-0.5em}
\end{figure}

\begin{figure}[t]
\centering
    \subfigure[Validation Accuracy]{
    \includegraphics[width=0.315\textwidth]{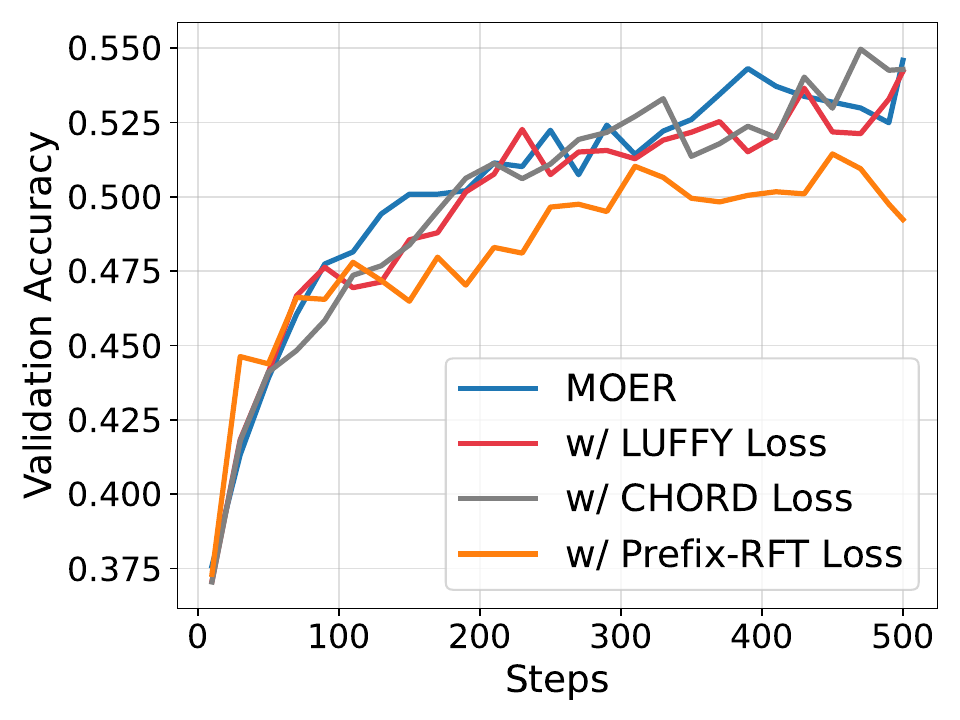}
    }
    \subfigure[Entropy]{
    \includegraphics[width=0.315\textwidth]{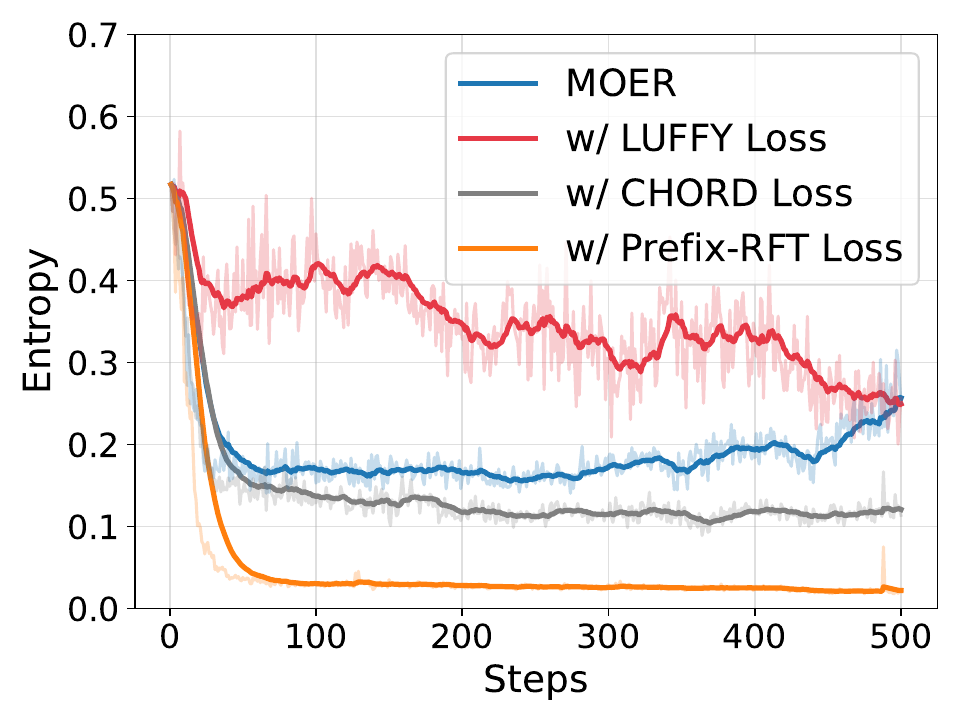}
    }
    \subfigure[Response Length]{
    \includegraphics[width=0.315\textwidth]{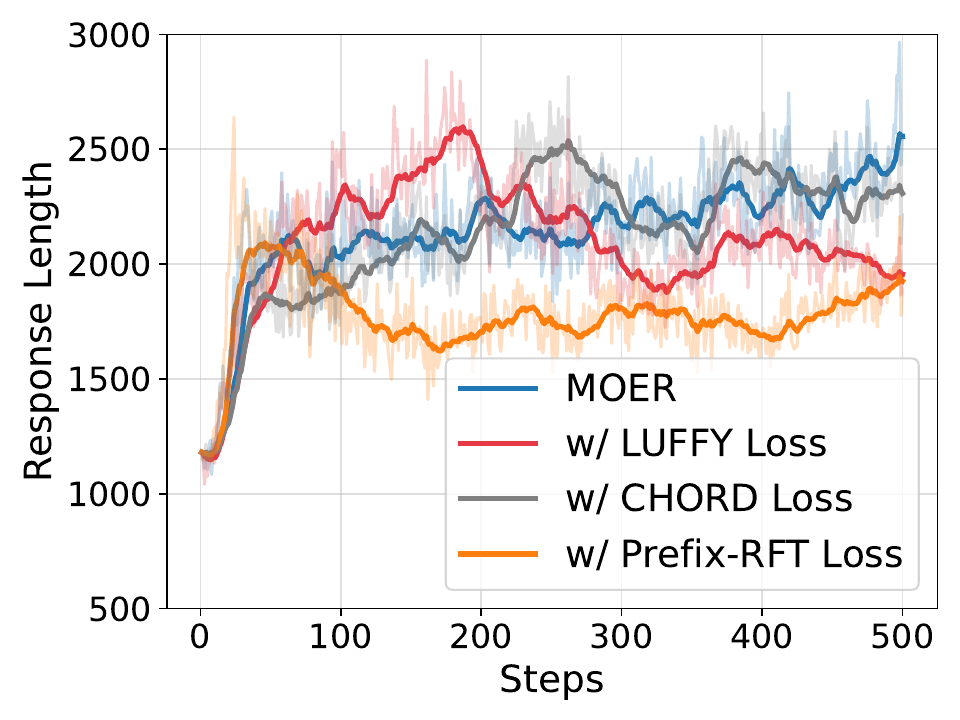}
    }
    \vspace{-0.5em}
    \caption{Training dynamics of different off-policy objectives.}
    \label{fig:analysis_off}
    \vspace{-0.5em}
\end{figure}

To empirically validate this, we substitute MOER's off-policy objective with those of other methods. As shown in Table~\ref{tab:off_obj_ab}, the scaled trust-region SFT objective adopted by MOER outperforms the alternatives. In addition, the training dynamics of different off-policy objectives are illustrated in Figure \ref{fig:analysis_off}. Specifically, Prefix-RFT objective leads to entropy collapse since the loss of most low-probability tokens are masked, while LUFFY objective induces high entropy due to the excessive gradient weights of these tokens. Although the training dynamic of the variant with CHORD objective is similar to that of MOER, its general domain performance is much worse than MOER. 

\section{Analysis on Clipping Operation for On-Policy Tokens} \label{app:analysis_clip}
\begin{figure}[h]
\centering
    \subfigure[Validation Accuracy]{
    \includegraphics[width=0.315\textwidth]{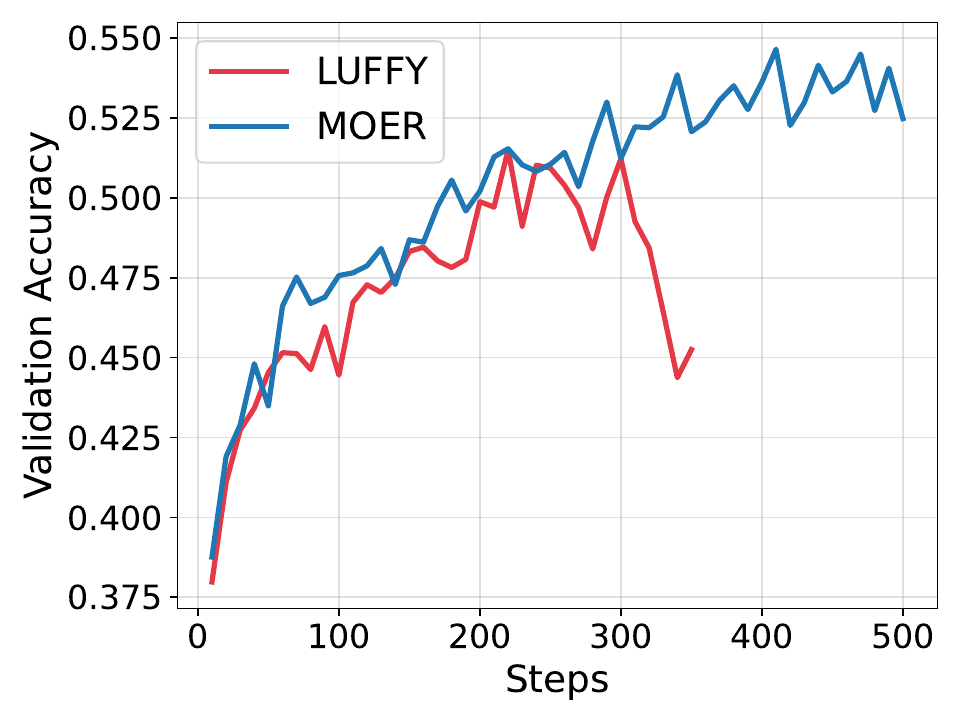}
    }
    \subfigure[Entropy]{
    \includegraphics[width=0.315\textwidth]{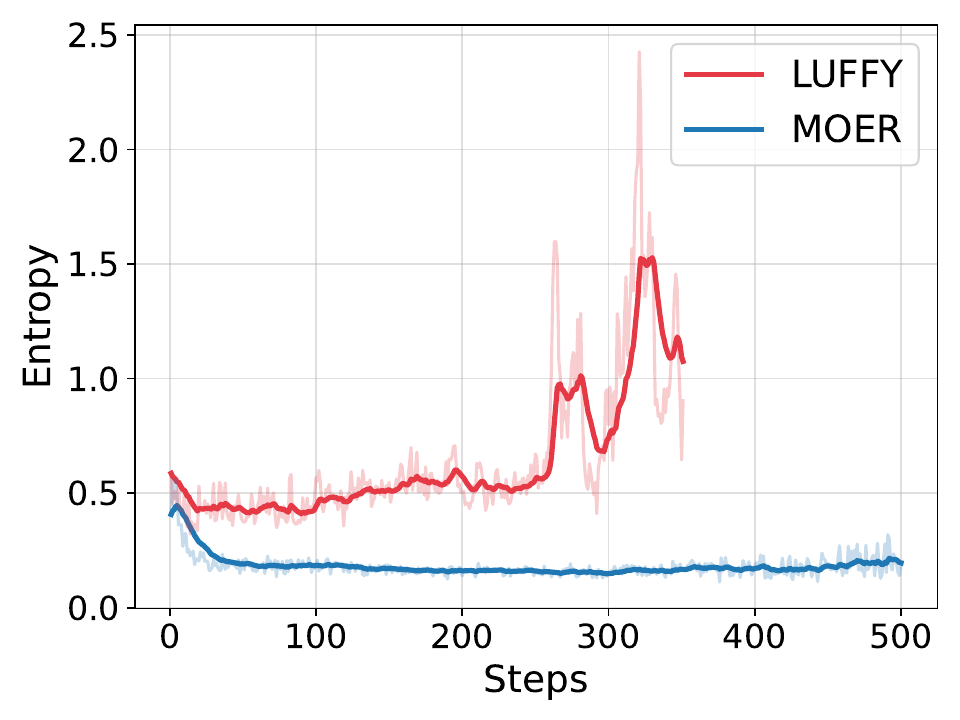}
    }
    \subfigure[Response Length]{
    \includegraphics[width=0.315\textwidth]{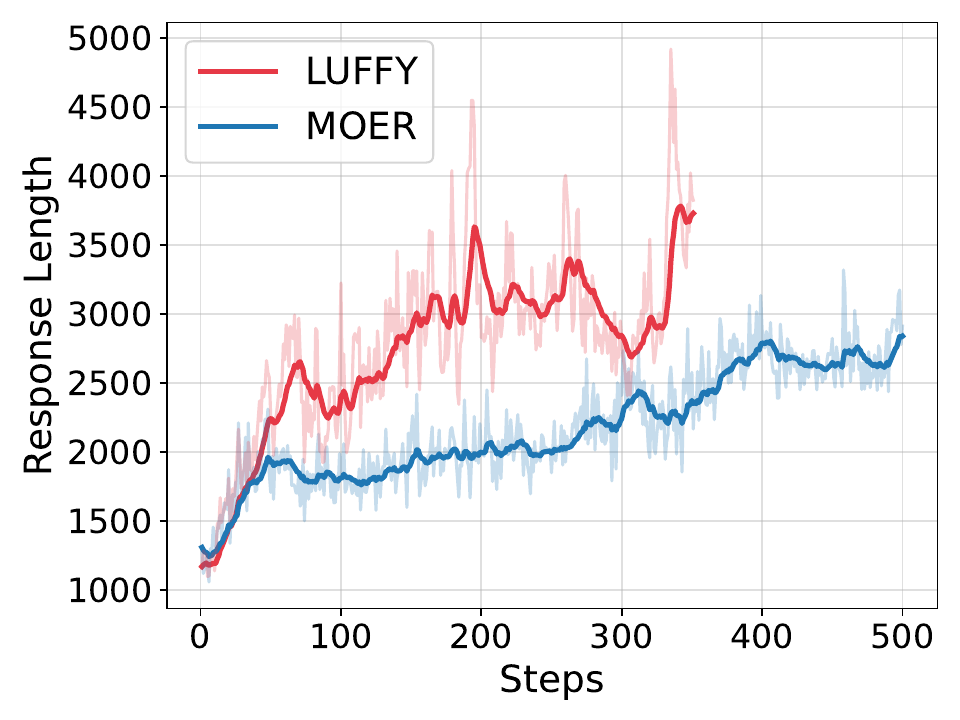}
    }
    \vspace{-0.5em}
    \caption{Training dynamics of LUFFY and MOER under different rollout/update batch size settings. When the rollout batch size is $64$ and the update batch size is $16$, i.e., the updates become more off-policy, LUFFY collapses in the middle of training, while MOER remains stable.}
    \label{fig:analysis_clip_bs}
    \vspace{-0.5em}
\end{figure}

The clipping operation on the importance sampling (IS) ratio serves as a trust-region safeguard in PPO \citep{schulman2017proximal} and GRPO \citep{shao2024deepseekmath}, preventing destructively large policy updates. LUFFY \citep{yan2026learning} removes it to allow greater flexibility in updating toward off-policy actions, but this exposes training to unbounded IS ratios. 

As shown in Figure \ref{fig:analysis_ab}, LUFFY already exhibits late-stage instability under the default setting, i.e., the rollout batch size is $128$ and the update batch size is $64$. This instability is substantially amplified when updates become more off-policy: as illustrated in Figure \ref{fig:analysis_clip_bs}, with rollout batch size $64$ and update batch size $16$, i.e., four gradient steps per rollout on increasingly stale data, LUFFY collapses in the middle of training with degrading performance and exploding entropy, since tokens whose probabilities have drifted from the rollout policy receive excessively large IS ratios and unbounded gradients, an effect that compounds with each additional gradient step. In contrast, MOER clips the IS ratio of on-policy tokens and remains stable under both settings. Nevertheless, as presented in Figure \ref{fig:ab_hp} (c) of Appendix \ref{app:ab_hp}, the clipping coefficients require careful tuning: the default symmetric coefficients of GRPO ($\epsilon_{low}=\epsilon_{high}=0.2$) cause entropy collapse, whereas our asymmetric coefficients ($\epsilon_{low}=0.2$, $\epsilon_{high}=0.24$) keep entropy in a stable range, neither collapsing nor exploding.

%% file: appendix/exp.tex
\section{Accuracy of Prefix-Conditioned Completions} \label{app:motivation}
\begin{figure}[h]
    \centering
    \includegraphics[width=0.58\textwidth]{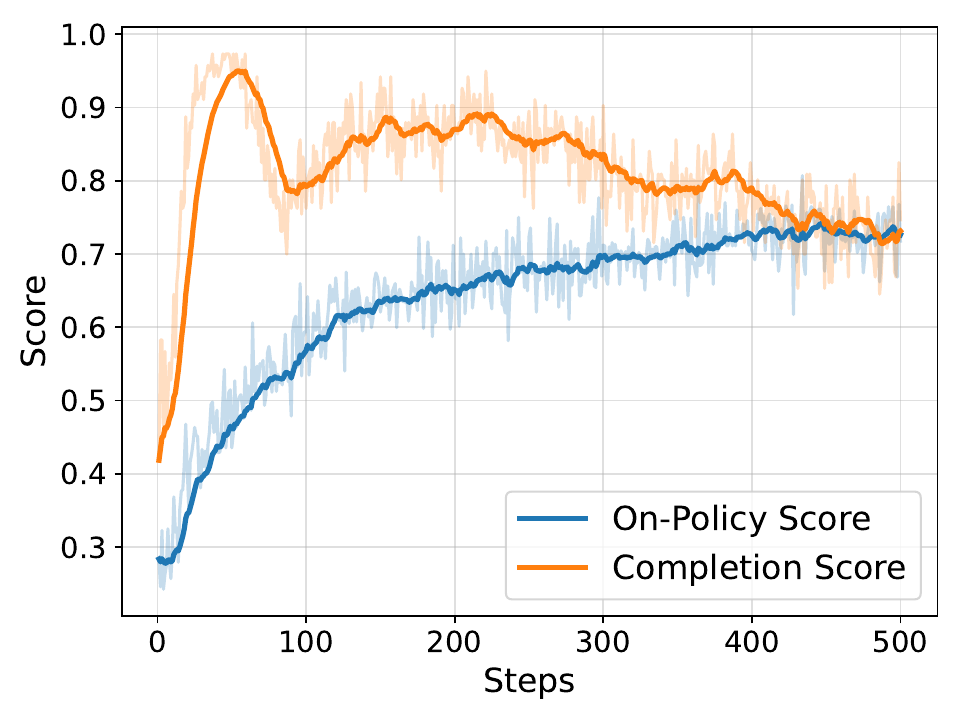}
    \vspace{-1em}
    \caption{Accuracy of on-policy rollouts and prefix-conditioned completions along training. Note that the first $50$ steps are warm-up stage, where the prefix is the full expert CoT.}
    \label{fig:comp_acc}
    \vspace{-0.5em}
\end{figure}

To examine how the benefit of expert prefix guidance evolves during training, we track the accuracy of prefix-conditioned completions together with that of pure on-policy rollouts in Figure \ref{fig:comp_acc}. Two observations stand out. First, prefix-conditioned completions consistently achieve higher accuracy than on-policy rollouts, confirming that expert prefixes provide effective guidance throughout the entire training process. Second, the gap between the two curves steadily closes: as the actual prefix ratio decays and the policy strengthens, the completion accuracy gradually converges towards the on-policy accuracy, indicating that the model progressively internalizes the reasoning ability that the prefix initially provides. Thus, the expert guidance diminishes over training, which supports the design of the decaying prefix schedule: guidance is most beneficial in early training and can be safely faded out as the policy's self-exploration capability strengthens. 

\section{Code Generation Experiments} \label{app:code}
\begin{table}[h]
    \centering
    \small
    \vspace{-1em}
    \caption{Results on the LiveCodeBench benchmark. We report the pass@1, entropy, and response length of MOER and LUFFY.}
    \label{tab:code}
    \vspace{0.5em}
    \begin{tabular}{l|ccc}
        \toprule
        Method & Pass@1 & Entropy & Response Length  \\
        \midrule
        LUFFY & 33.8 & 0.453 & 1903 \\
        \textbf{MOER}  & \textbf{38.3} & 0.246 & 668 \\
        \bottomrule
    \end{tabular}
\end{table}
We use CodeForces-CoTs \citep{penedo2025codeforces} as the training set. During preprocessing, we filter out samples whose prompt and response lengths exceed $1024$ and $8192$ tokens, respectively, and split the remaining data into $13000$ training samples and $1402$ validation samples. We use the public test cases to verify the correctness of the generated code, and remove these test cases from the input prompts to prevent reward hacking. We adopt LiveCodeBench-CPP \citep{jain2025livecodebench} as the evaluation benchmark and Qwen2.5-Instruct-7B as the base model.

We compare MOER with LUFFY, increasing LUFFY's policy shaping coefficient $\gamma$ from $0.1$ to $0.2$ to stabilize its training. We train the models for $380$ steps. All other hyperparameters follow Sec. \ref{sec:exp_setting}.

As shown in Table \ref{tab:code}, MOER achieves substantially stronger coding performance than LUFFY, while maintaining smaller yet healthy entropy and producing much shorter responses. These results indicate that the gains of MOER generalize beyond mathematical reasoning to code generation tasks.

\section{Results on Larger Model} \label{app:14b}
\begin{table*}[t]
\centering
\setlength{\tabcolsep}{4pt}
\vspace{-1em}
\caption{Experiment results on mathematical reasoning benchmarks based on Qwen2.5-Instruct-14B.}
\label{tab:14b}
\vspace{0.5em}
\small
\begin{tabular}{l ccccc >{\columncolor{yellowavg}}c}
\toprule
\multirow{2}{*}{\textbf{Method}} & \multicolumn{6}{c}{\textbf{Math Reasoning Performance}}\\
\cmidrule(lr){2-7} 
 & \textbf{AIME 24/25} & \textbf{AMC} & \textbf{MATH-500} & \textbf{Minerva} & \textbf{Olympiad} & \textbf{Avg.}  \\
\midrule
\multicolumn{7}{c}{\textit{Qwen2.5-Instruct-14B}} \\
\midrule
LUFFY &  18.4/\textbf{18.5} & 52.9 & 84.4 & \textbf{46.3} & 48.3 & 44.8 \\
\textbf{MOER} & \textbf{25.7}/14.3 & \textbf{54.9} & \textbf{85.2} & 43.0 & \textbf{52.0} & \textbf{45.9} \\
\bottomrule
\end{tabular}
\vspace{-0.5em}
\end{table*}

To further validate the generalizability of our method on larger models, we evaluate MOER and LUFFY on Qwen2.5-Instruct-14B \citep{qwen2025qwen25technicalreport}. Note that the results of ReLIFT are not included due to unavailable checkpoint and prohibitive computational costs.
As presented in Table \ref{tab:14b}, our method outperforms LUFFY across most mathematical reasoning tasks. Notably, MOER yields a more pronounced performance gain on 14B models compared to its 1.5B and 7B counterparts. Notably, while LUFFY leads to training collapse in the middle of training, MOER still guarantees training stability throughout the entire optimization process.

\section{Ablation Study on Hyperparameters} \label{app:ab_hp}
\begin{figure}[h]
\centering
\vspace{-0.5em}
    \subfigure[Minimum Prefix Ratio $r_{min}$]{
    \includegraphics[width=0.315\textwidth]{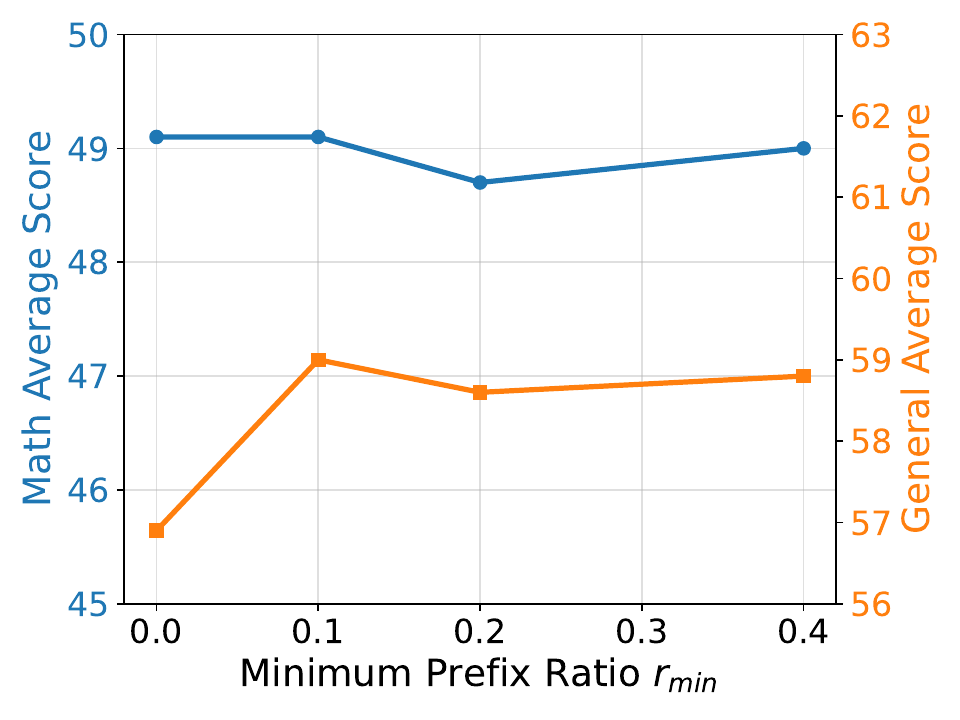}
    }
    \subfigure[Off-Policy Loss Weight $\alpha$]{
    \includegraphics[width=0.315\textwidth]{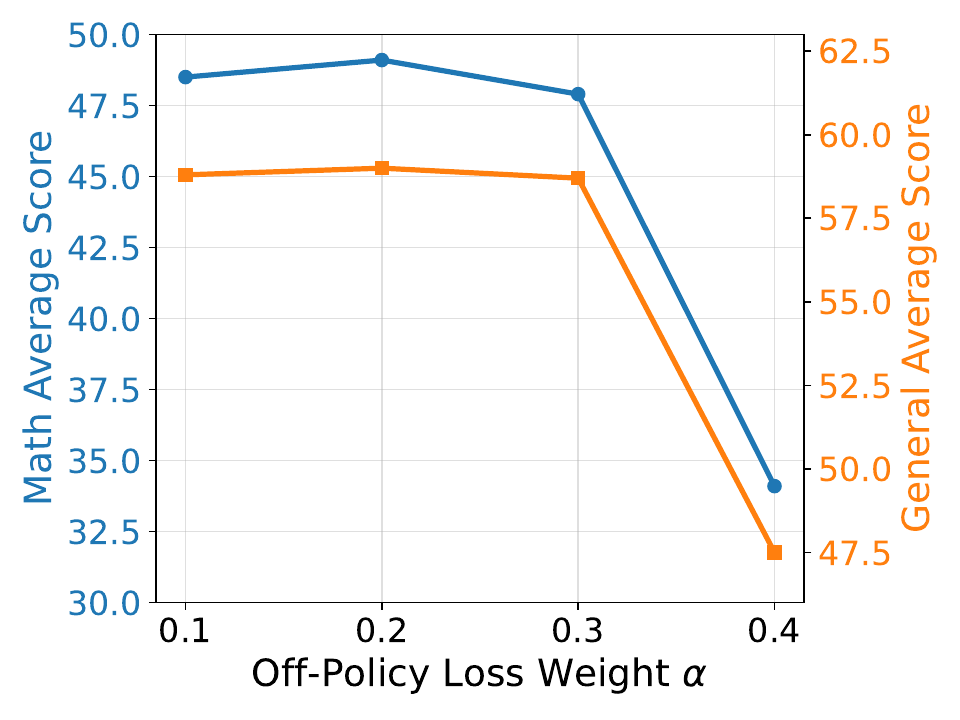}
    }
    \subfigure[Clipping Coefficient $\epsilon_{high}$]{
    \includegraphics[width=0.315\textwidth]{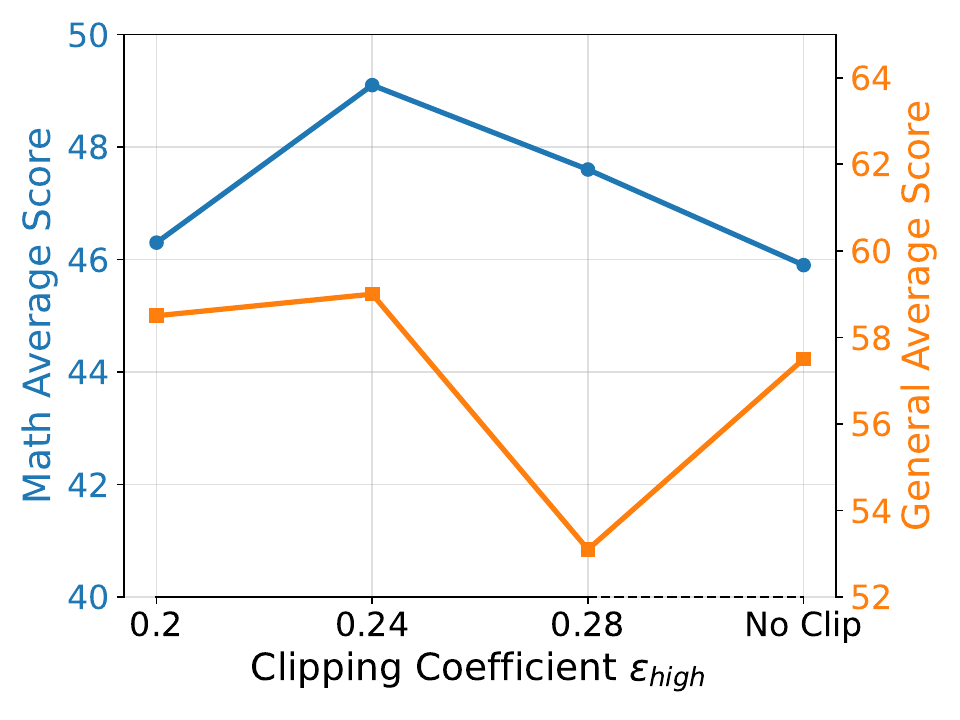}
    }
    \vspace{-0.5em}
    \caption{Ablation study on hyperparameters of MOER. The blue and orange lines denote the math and general domain average scores, respectively.}
    \label{fig:ab_hp}
    \vspace{-0.5em}
\end{figure}

In this section, we conduct an ablation study on some critical hyperparameters in MOER, including the minimum prefix ratio $r_{min}$, off-policy loss weight $\alpha$, and clipping coefficient $\epsilon_{high}$. Note that the experimental setting is consistent with Sec. \ref{sec:ab}.

\textbf{Minimum prefix ratio $\bm{r_{min}}$.}
As shown in Figure \ref{fig:ab_hp} (a), the performance on both math and general domain benchmarks varies only marginally across different values of $r_{min}$, indicating that MOER is relatively robust to the minimum prefix ratio. We set $r_{min}=0.1$ by default.

\textbf{Off-policy loss weight $\bm{\alpha}$.}
As shown in Figure \ref{fig:ab_hp} (b), $\alpha=0.2$ achieves the best overall performance. A smaller weight under-utilizes expert guidance, while an excessively large weight strengthens off-policy supervision too aggressively and degrades performance.

\textbf{Clipping coefficient $\bm{\epsilon_{high}}$.}
As shown in Figure \ref{fig:ab_hp} (c), $\epsilon_{high}=0.24$ yields the best performance. Notably, since $\epsilon_{low}$ is fixed at $0.2$, setting $\epsilon_{high}=0.2$ degenerates to the symmetric clipping of standard GRPO, which leads to entropy collapse and finally degraded response diversity. Conversely, an overly large $\epsilon_{high}$, or removing the clipping operation entirely, causes instability and ultimately suboptimal performance, which is consistent with our analysis in Appendix \ref{app:analysis_clip}.

\section{Pass@k Results} \label{app:pass}
\begin{figure}[h]
\centering
    \includegraphics[width=0.65\textwidth]{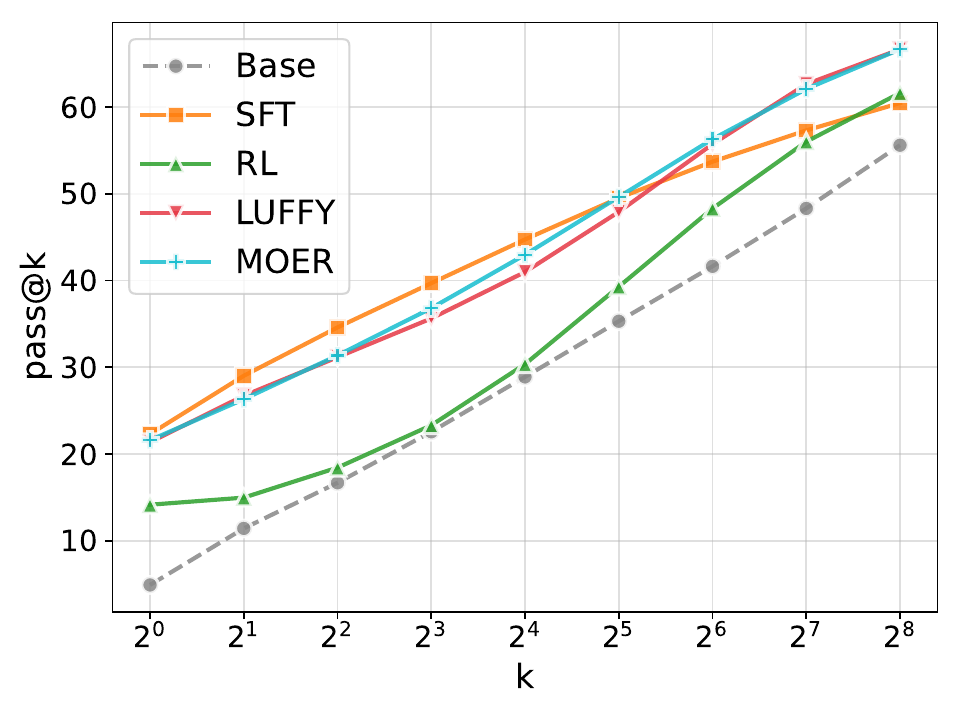}
    \vspace{-1em}
    \caption{Pass@k results of MOER and baselines on the AIME 2025 benchmark. The base model is Qwen2.5-Math-7B. We report the results with $k=2^0, 2^1,..., 2^8$.}
    \label{fig:pass}
\end{figure}

To verify whether MOER incentivizes the model's intrinsic reasoning capacity rather than merely increasing the sampling probability of knowledge already present in the base model, we evaluate our method under varying computational budgets. Specifically, we report the pass@k results ($k \in \{2^0, 2^1, \dots, 2^8\}$) of the Qwen2.5-Math-7B base model, SFT, RL (GRPO), LUFFY, and MOER on the challenging AIME 2025 benchmark in Figure~\ref{fig:pass}.

The results indicate that both mix-policy methods, MOER and LUFFY, consistently outperform the base model and standard RL across all sampling budgets ($k$). In the low-budget regime ($k \le 2^4$), SFT achieves slightly higher accuracy than mix-policy methods due to its concentrated probability mass on standard solution trajectories. However, as the computational budget increases ($k \ge 2^5$), the scaling curves of MOER and LUFFY cross that of SFT and continue to ascend sharply, reaching $66.7\%$ pass@$2^8$ performance compared to SFT's $60.5\%$. This crossover demonstrates that MOER genuinely enhances the model's underlying problem-solving capability and solution diversity rather than overfitting to superficial response patterns. 

Furthermore, while MOER exhibits scaling performance similar to LUFFY on this mathematical benchmark, it achieves this parity with notably superior general-domain utility, shorter response lengths, and better training stability (as detailed in Sec.~\ref{sec:ab}). This confirms that MOER maintains high reasoning efficiency and generalizability without incurring superfluous generation overhead during sampling.